\documentclass{article}

 \usepackage[preprint]{neurips_2026}

\usepackage[utf8]{inputenc} 
\usepackage[T1]{fontenc}    
\usepackage{hyperref}       
\usepackage{url}            
\usepackage{booktabs}       
\usepackage{amsfonts}       
\usepackage{nicefrac}       
\usepackage{microtype}      
\usepackage{xcolor}         
\usepackage{wrapfig}

 \usepackage{multirow}
 \usepackage{amsmath}
 \usepackage{algorithm}
 \usepackage{algpseudocode}
 \usepackage{float} 
 \usepackage{graphicx} 
 \usepackage{enumitem}
 \usepackage{makecell}
 \usepackage{caption}
 \usepackage{wrapfig}
\usepackage{amssymb}
\usepackage{amsthm}

\usepackage{mathrsfs}

\newtheorem{proposition}{Proposition}

\definecolor{DarkGreen}{rgb}{0.0, 0.0, 0.5}
\title{A New Perspective on ANN-to-SNN Conversion: Hardware-Aligned Multi-Threshold Neuron Dynamics for Neuromorphic Inference}

\author{
  Shaohong Gu\textsuperscript{1,2} \\
  Nanhu Academy of Electronics and Information Technology \\
  College of Electronic Engineering and Information Science \\
  Zhejiang University \\
  Jiaxing, Zhejiang / Hangzhou, Zhejiang, China \\
  \And
  Jiyuan Zhang\textsuperscript{3} \\
  School of Computer Science and Engineering \\
  University of Electronic Science and Technology of China \\
  Chengdu, Sichuan, China \\
  \And
  Yimeng Shan\textsuperscript{3} \\
  School of Computer Science and Engineering \\
  University of Electronic Science and Technology of China \\
  Chengdu, Sichuan, China \\
  \And
  Chenxiao Dou\textsuperscript{1,*} \\
  Nanhu Academy of Electronics and Information Technology \\
  Jiaxing, Zhejiang, China \\
  \texttt{douchenxiao@cnaeit.com} \\
  \And
  Wei Hua\textsuperscript{1} \\
 Nanhu Academy of Electronics and Information Technology \\
  Jiaxing, Zhejiang, China \\
  \And
  Malu Zhang\textsuperscript{3} \\
  School of Computer Science and Engineering \\
  University of Electronic Science and Technology of China \\
  Chengdu, Sichuan, China \\
  \And
  Honglin Cao\textsuperscript{3} \\
  School of Computer Science and Engineering \\
  University of Electronic Science and Technology of China \\
  Chengdu, Sichuan, China \\
  \And
  Choonghyun Lee\textsuperscript{2} \\
  College of Electronic Engineering and Information Science \\
  Zhejiang University \\
  Hangzhou, Zhejiang, China \\
  \And
  Jingwei Chen\textsuperscript{1} \\
  Nanhu Academy of Electronics and Information Technology \\
  Jiaxing, Zhejiang, China \\
}

\begin{document}

\maketitle

\begin{abstract}

By accumulating spikes over multiple timesteps, rate coding approximates the real-valued activations of an ANN, a principle that underpins numerous ANN-to-SNN conversion methods. However, the converted SNNs typically incur loss in accuracy compared to the original ANN. 
Existing methods either increase a number of timesteps or introduce more complex neuronal dynamics to reduce this accuracy gap. Yet, both approaches ultimately lead to the same outcome: redundancy either in information or in computation, thereby suffering from efficiency degradation. %
To address this issue, we propose \textbf{PRIME} (\underline{P}otential-\underline{R}esidual \underline{I}ntegration for \underline{M}ulti-threshold \underline{E}ncoding), a multi-threshold concurrent spiking neuron framework redesigned from a hardware data representation perspective:
a) We introduce the Residual Membrane Potential Amplification (RMPA) mechanism, which replaces rate coding with Integration Coding by actively amplifying the residual membrane potential. RMPA enables the binary representation of spikes to carry substantially more information, achieving near-lossless conversion with as few as two timesteps.
b) Complementarily, we implement concurrent spike firing in multi-threshold neurons, where spikes are decoded directly from the binary representation of floating-point data, eliminating membrane potential--threshold comparisons and replacing multiplications with lightweight bit-shift and accumulation operations that align with hardware computation primitives. At $T=2$ and $D=3$, we achieve competitive performance across both vision and language benchmarks with only 40.6\% of the FP32 bit-transmission cost and 13.4\% of the FP32 multiplication energy consumption. The PRIME framework establishes a new design principle for efficient ANN-to-SNN conversion, bridging the gap between algorithmic efficiency and hardware constraints.

\end{abstract}

\section{Introduction}

 Human brains execute high-level cognitive tasks at remarkably low energy cost~\cite{gerstner2014neuronal,izhikevich2003simple}, offering a compelling biological blueprint for energy-efficient hardware design. Spiking neural networks (SNNs) emulate spike-based biological communication, inheriting high sparsity and event-driven processing~\cite{maass1997networks,davies2018loihi}. While direct training via surrogate gradients suffers from approximation errors and expensive spatio-temporal backpropagation~\cite{gygax2024,zhu2024osr}, ANN-to-SNN conversion has emerged as the dominant cost-effective paradigm for obtaining high-performance SNNs~\cite{cao2015spiking}.

Existing conversion methods rely predominantly on rate coding~\cite{rueckauer2017conversion} and fall into two categories: (1) \textbf{extending timesteps}, where fidelity asymptotically improves with more timesteps~\cite{jelly}, and (2) \textbf{introducing auxiliary neuronal dynamics}, such as differential encoding or multi-threshold mechanisms that require storing multiple pre-set thresholds~\cite{dcgs,spikellm}. The former incurs \textit{information redundancy}, as a single activation value requires sizable spikes across many timesteps; the latter introduces \textit{computational redundancy} through complex computations imposed by auxiliary neuronal dynamics. Both pathways share a common root: neither exploits the intrinsic alignment between SNN spike representations and the binary data formats native to modern hardware.

We identify two complementary insights that address this gap. 
The conversion error stems from residual membrane potential — the unencoded input remaining after all timesteps. Conventional approaches passively wait for this residual to dissipate, yielding only marginal per-step gains. Actively amplifying the residual instead enables exponential suppression of conversion error. From a hardware perspective, binary spike sequences share the same underlying representation as stored binary data, suggesting that spike thresholds and amplitudes can be read directly from floating-point bit fields, tightly coupling neuronal dynamics with the native hardware data format.

Guided by these insights, we propose \textbf{PRIME} (\underline{P}otential-\underline{R}esidual \underline{I}ntegration for \underline{M}ulti-threshold \underline{E}ncoding). \textbf{At the encoding level}, the Residual Membrane Potential Amplification (RMPA) mechanism actively amplifies residual membrane potential per step, enabling near-lossless conversion at $T \leq 2$ without auxiliary compensation. \textbf{At the inference level}, spike thresholds and amplitudes are read directly from the sign, exponent, and mantissa bits of floating-point values, eliminating membrane potential accumulation and multi-threshold comparison. Reading $D$ consecutive mantissa bits enables concurrent multi-bit spike emission per step; when $T \leq 2$, inter-step independence allows fully parallel inference with multiplications replaced by shift-and-add operations. PRIME simultaneously resolves the information redundancy of time-step extension and the computational redundancy of threshold storage. Extensive experiments on CV and NLP tasks validate its effectiveness. Our main contributions are summarized as follows:
\begin{itemize}

\item \textbf{At the encoding level}, we propose the RMPA mechanism and Integration Coding, which actively amplify residual membrane potential to maximize per-step information density, achieving near-lossless conversion at $T \leq 2$ without auxiliary compensation.

\item \textbf{At the inference level}, we propose a hardware-native spike-driven inference scheme that couples neuronal dynamics with floating-point bit representations, replacing multiplications with shift-and-add operations. When $T \leq 2$, inter-step independence enables fully parallel inference; when $D \leq 3$, each intra-step computation requires at most a single adder.

\item Building upon these two designs, we propose \textbf{PRIME}, a unified ANN-to-SNN conversion framework that jointly addresses encoding inefficiency and hardware deployment overhead.

\item Extensive experiments demonstrate that under $T=2$, $D=3$, PRIME achieves near-lossless conversion accuracy on representative CV and NLP benchmarks, with energy consumption at 13.4\% of FP32 multiplication and bit-transmission cost between layers at 40.6\% of FP32, providing a strong foundation for energy-efficient neuromorphic hardware design.
\end{itemize}



\section{Related Work}

\subsection{Training-Dependent Conversion}
 
Training-dependent methods retrain the source ANN to align activations with spiking neuron dynamics before conversion. Under rate coding, representative works optimize membrane potential initialization~\cite{bu2022optimized}, replace ReLU with quantization-aware activations~\cite{bu2023optimal}, or apply dual-phase pipelines for weight and threshold co-calibration~\cite{ding2021optimal, wang2022towards}. For Transformers, SpikeZIP-TF~\cite{you2024spikezip} fine-tunes pretrained weights with spike-compatible surrogates, and SpikeLLM~\cite{xing2024spikellm} extends spiking mechanisms to billion-parameter LLMs via saliency-aware encoding. While effective, these methods require training pipeline access, limiting applicability when the source ANN is fixed or proprietary.

\subsection{Training-Free Conversion}
 
In contrast, Training-free approaches convert pretrained ANNs directly without fine-tuning. The dominant paradigm is rate coding~\cite{rueckauer2017conversion, diehl2015fast}, which requires at least $T = N$ timesteps to distinguish $N$ activation levels. To mitigate this, prior works calibrate residual and initial membrane potentials~\cite{hao2023reducing, hao2023bridging}, extend neuron outputs to ternary values~\cite{wang2022signed}, or refine threshold calibration strategies~\cite{wang2025adaptive, bu2025inference}. Beyond rate coding, phase coding~\cite{kim2018deep, wang2022efficient} supports $2^T$ representable values but requires all timesteps to elapse before decoding, while differential coding (DCGS)~\cite{huang2025differential} eliminates temporal decay at the cost of hardware-unfriendly division operations. For Transformer conversion, ECMT~\cite{huang2024towards} employs multi-threshold neurons but retains floating-point multiplication; STA~\cite{jiang2024spatio} approximates arbitrary nonlinearities but requires hundreds of timesteps; and SpikedAttention~\cite{hwang2024spikedattention} achieves fully spike-driven conversion via trace-driven matrix multiplication. 
Despite these advances, all existing methods rely on iterative accumulation and gradual dissipation of residual membrane potential, encoding redundant information across timesteps — a fundamental bottleneck that neither algorithmic nor hardware optimizations have yet resolved.
\section{Method}

In this section, we propose the \textbf{P}otential-\textbf{R}esidual \textbf{I}ntegration for \textbf{M}ulti-threshold \textbf{E}ncoding (\textbf{PRIME}) framework, which optimizes the multi-threshold paradigm along two complementary dimensions: a more expressive \emph{encoding strategy} and an \emph{inference-oriented hardware design} tailored to it.

At the encoding level, PRIME introduces Integration Coding together with its companion Residual Membrane Potential Amplification (RMPA) mechanism, which progressively doubles the residual membrane potential so that MTNs deterministically recover previously unencoded information, achieving rapid convergence toward the corresponding ANN activation values.

At the inference level, PRIME co-designs spike generation and multiplication around the bit-level structure of FP16 operands: Direct FP16 Binary Parsing maps the sign, exponent, and mantissa fields directly to thresholds and spikes via a leading-one detector, while spike-driven multiplication reduces the mantissa product to trace-aligned shift-add operations, eliminating the usage of multipliers.


\subsection{Preliminary}
\label{sec:pre}


 
Following the successful practice of many prior works~\cite{you2024spikezip,wang2022signed,huang2025differential,huang2024towards}, in this paper, we adopt ternary-spike-based MTNs with outputs in $\{-1, 0, +1\}$ as the core computational unit of the SNN. Each MTN in layer $\ell$ is characterized by a base threshold $\theta^\ell$ and $2n$ thresholds, comprising $n$ positive and $n$ negative ones. The $i$-th threshold in layer $\ell$, denoted $\lambda_i^\ell$, is defined as $\lambda_i^\ell = \theta^\ell / 2^{i-1}$ for the positive branch ($i = 1, \ldots, n$) and $\lambda_{n+i}^\ell = -\theta^\ell / 2^{i-1}$ for the negative branch ($i = 1, \ldots, n$), forming a symmetric set of geometrically decaying values around zero.

Most ANN-to-SNN conversion methods are built upon rate coding~\cite{you2024spikezip,wang2022signed,wang2026training}. The neuronal dynamics of rate-coding-based MTNs can be described as the following equations: 
\begin{equation}
\begin{aligned}
\mathbf{m}^\ell[t] &= \mathbf{V}^\ell[t-1] + \mathbf{I}^{\ell}[t], &\quad
\mathbf{S}_i^\ell[t] &= \mathrm{MTH}_{\theta,n}(\mathbf{m}^\ell[t], i), \\
\mathbf{X}^\ell[t] &= \sum_i \mathbf{S}_i^\ell[t]\,\mathbf{W}^\ell \lambda_i^\ell, &\quad
\mathbf{V}^\ell[t] &= \mathbf{m}^\ell[t] - \mathbf{X}^\ell[t],
\end{aligned}
\end{equation}
\begin{equation}
\mathrm{MTH}_{\theta,n}(\mathbf{m}^\ell[t], i) = 
\begin{cases}
1, & \text{if } i = i^*(t) \text{ and } \mathbf{m}^\ell[t] \notin (\lambda_{2n}, \lambda_n), \\
0, & \text{otherwise},
\end{cases}
\end{equation}

where $\mathbf{m}^\ell[t]$, $\mathbf{V}^\ell[t]$, $\mathbf{I}^\ell[t]$, $\mathbf{S}_i^\ell[t]$, 
$\mathbf{W}^\ell$ and $\mathbf{X}^\ell[t]$ denote the pre-spike membrane potential, 
post-spike membrane potential, input current, output spike of the $i$-th 
threshold, weight, and total output signal of layer $\ell$ at timestep $t$, 
respectively. Here $i^*(t) \triangleq \arg\min_{p \in \{1,\ldots,2n\}} 
|\mathbf{m}^\ell[t] - \lambda_p|$ denotes the index of the threshold closest to the 
current membrane potential, so that the MTN fires a spike on the $i$-th 
channel only when $i$ matches this nearest threshold and $\mathbf{m}^\ell[t]$ lies 
outside the dead zone $(\lambda_{2n}, \lambda_n)$. Under rate coding, the 
quantization error between the ANN activation and the SNN output arises 
from the residual membrane potential that remains below the firing 
threshold. Existing methods mitigate this error by increasing the number 
of timesteps or thresholds, both of which incur additional memory, 
computation, and energy overhead, thereby impeding edge deployment. To 
address this, we propose integration coding-based MTNs, which directly 
exploit the residual membrane potential without requiring extensive 
time-step iterations or a large number of thresholds.

\subsection{Multi-Threshold Neuron Based on Integration Coding}

\begin{figure}[htbp]
    \centering
    \includegraphics[width=1.0\linewidth]{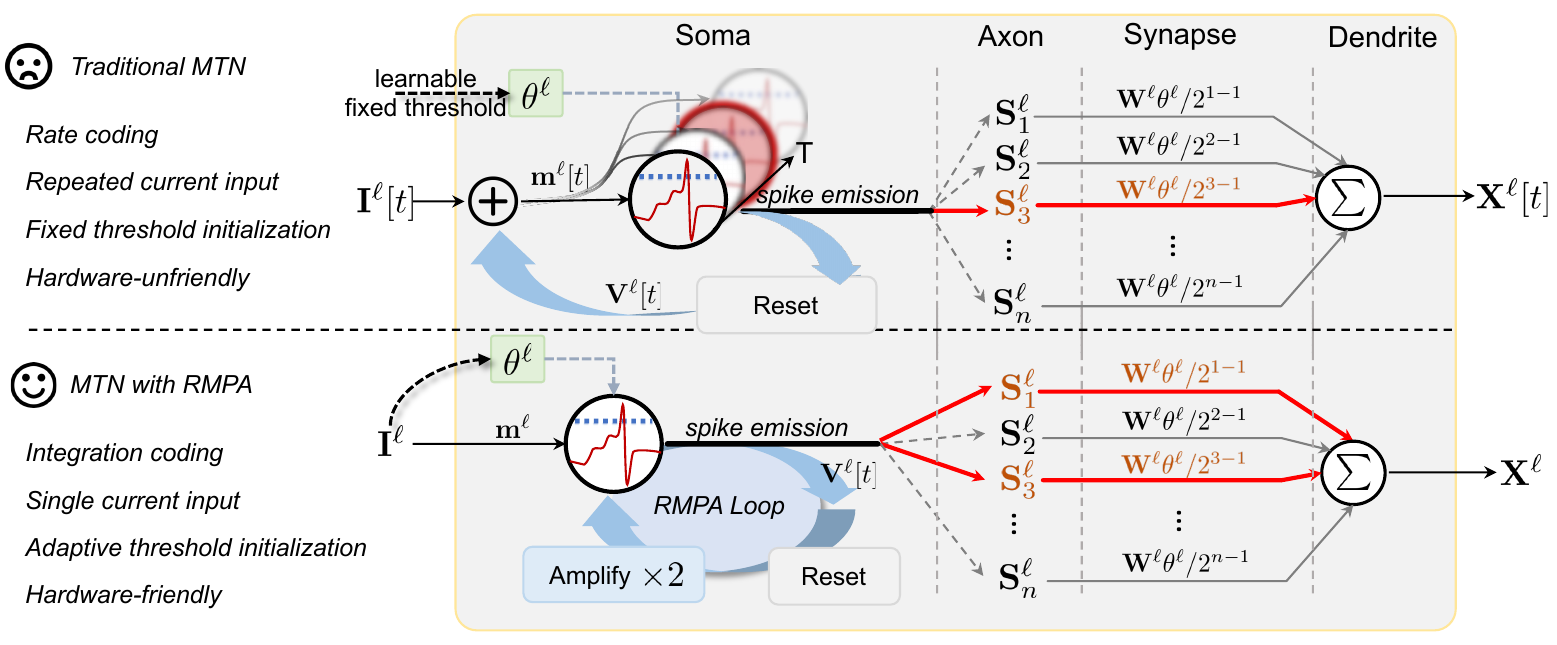}
    \caption{Comparison between the traditional MTN and the proposed MTN with RMPA. \textbf{Upper}: traditional MTN. Repeated current input introduces redundant addition operations, while fixed threshold initialization requires the membrane potential to be compared with multiple thresholds, resulting in extra hardware computation overhead. \textbf{Lower}: the MTN with RMPA, which uses a single current input and adaptive threshold initialization for more hardware-friendly implementation. }
    \label{Integral_Coding}
\end{figure}

The dynamics of the integration coding-based MTN are illustrated in Figure~\ref{Integral_Coding}. Conventional rate coding-based neurons accumulate input current at every timestep, whereas the integration coding-based MTN receives input only once, so the input current $\mathbf{I}^\ell$ directly initializes the membrane potential $m^\ell = \mathbf{I}^\ell$.
Afterwards, the neuron identifies the largest 
threshold that triggers spike emission, aligned with the binary structure of floating-point numbers 
(detailed in the Subsection~\ref{subsec:spike-gen-via-fp16-binary-parsing}). At the beginning, the base threshold $\theta^\ell$ is set and followed by a soft reset according to Eq.~\eqref{eqa:thres}:
\begin{equation}
\label{eqa:thres}
\theta^\ell = 2^p \text{ where } p = \max\{p \mid 2^p \leq \mathbf{m}^\ell,\ p \in \mathbb{Z}\}, \quad\mathbf{V}^\ell = \mathbf{m}^\ell - \frac{\mathbf{S}_i^\ell \theta^\ell}{2^{i-1}},
\end{equation}
where $i$ indicates the loop index of RMPA.
In each subsequent iteration, $\mathbf{m}^\ell$ is set to the post-reset membrane potential $\mathbf{V}^\ell$.

The RMPA-based MTN iteratively generates sub-threshold spikes from the residual membrane potential, progressively reducing the quantization error. Each spike generation requires only a bit-shift and a single comparison against the base threshold $\theta ^\ell$, eliminating storing a large number of thresholds. Formally, the firing rule is defined as Eq.~\eqref{eqa:spike}:
\begin{equation}
\label{eqa:spike}
\mathbf{S}_i^\ell = \mathrm{MTH}_{\theta,n}(\mathbf{m}^\ell, i) = 
\begin{cases}
1, & \text{if } \mathbf{m}^\ell \cdot 2^{i - 1} - {\theta ^\ell} \ge 0\, \\
0, & \text{otherwise}
\end{cases}.
\end{equation}

A soft reset is then applied as defined in Eq.~\eqref{eqa:thres}. Once the membrane potential reaches zero or the prescribed number of iterations is exhausted, the neuron decodes the emitted spike sequence to produce the total output signal of layer $\ell$:
\begin{equation}
{\mathbf{X}}^{\ell} = \sum_{i}\mathbf{S}_i^\ell \mathbf{W}^\ell \frac{ \theta^\ell}{2^{i-1}}.
\end{equation}

We illustrate the internal dynamics of RMPA with a concrete example. Consider an input $x = 1.4$. According to Eq.~\eqref{eqa:thres}, the threshold $\theta^\ell$ is set to $1$ ($2^{p}$, $p=0$). After a spike of $\theta^\ell=1$ is emitted, it leaves $\mathbf{m}^\ell=1.4-1$ as $0.4$ and the index $i$ of RMPA loop as $1$. 
In the second iteration, $\mathbf{m}^\ell$ is doubled to $0.8$ and $i$ advances to $2$; since $\mathbf{m}^\ell = 0.8 < \theta^\ell=1$, no spike is emitted. 
In the third iteration, $\mathbf{m}^\ell$ is doubled to $1.6$ and $i$ advances to $3$; since $\mathbf{m}^\ell = 1.6 > \theta^\ell=1$, a spike of $\theta^\ell / 2^{i-1}= 0.25$ is fired and $\mathbf{m}^\ell$ is reset to $1.6 - 1 = 0.6$. 
At this point, the total encoded value is $1 + 0.25 = 1.25$, with residual membrane potential $\mathbf{m}^\ell = 0.6$. The remaining encoding error is $1.4 - 1.25$, the same as the result of $\mathbf{m}^l / 2^{i-1} = 0.6 / 4$, which decays exponentially with increasing iterations.


Our neurons offer two key advantages: (i) threshold storage is reduced from $2n$ to $\mathcal{O}(1)$, with a corresponding reduction in I/O operations; (ii) unlike rate coding, integration coding-based MTNs actively exploit the residual membrane potential to suppress quantization error, which is the fundamental source of their accuracy gain.

\begin{proposition}[Error Bound of the RMPA-based MTN]
\label{prop:rmpa-mtn-error-bound}
    The encoding error of conventional rate coding is bounded by
The encoding error of conventional rate coding is bounded by
    $\varepsilon_{\text{rate}} < \theta/2T,
    \label{eq:bound_rate}$
which decreases only linearly with $T$. In contrast, the proposed PRIME encoding achieves an exponentially tighter bound of
$\varepsilon_{\text{PRIME}} < \theta/2^{T \cdot D + 1},
    \label{eq:bound_prime}$
where $\theta$ denotes the spike threshold, $T$ the number of inter-timesteps, and $D$ the number of concurrent intra-timestep spikes.
\end{proposition}
Proof can be found in Appendix~\ref{sec:error}. Therefore, our proposed method can reduce several orders of magnitude compared to conventional MTN: for instance, at $T=2$ and $D=3$, the bound of PRIME is $\theta / 2^{7} = \theta / 128$, which is $32\times$ tighter than the rate coding bound of $\theta / 2T = \theta / 4$ at the same $T$.

\subsection{Hardware-efficient Inference}
\subsubsection{Spike Generation via Direct FP16 Binary Parsing}
\label{subsec:spike-gen-via-fp16-binary-parsing}
Here we describe how the spike generation process is directly bound to the binary representation of FP16 data, eliminating conventional membrane potential computation. The key insight is that each field of the FP16 format: sign, exponent, and mantissa can be mapped to the threshold and the spike corresponding to Eq.~\eqref{eqa:thres} and ~\eqref{eqa:spike}, enabling a purely read-based spike emission mechanism.

A normalized FP16 floating-point number is represented as:
\begin{equation}
    x_{\textbf{10}} = (-1)^{S} \times 2^{E-15} \times \left(1 + \sum_{j=1}^{10} M_j \times 2^{-j}\right),
\end{equation}
where $S$ is the 1-bit sign, $E$ is the 5-bit exponent ($1 \leq E \leq 30$), and $M_1 M_2 \cdots M_{10}$ is the 10-bit mantissa with an implicit leading 1. 
In the first cycle, the implicit leading 1 of the FP16 representation is directly emitted as the first spike, and the sign bit $S$ together with the exponent $E$ determines the corresponding threshold $\theta_1 = (-1)^{S} \times 2^{E-15}$. Starting from the second cycle, the position of the first 1 in the remaining mantissa is located using a leading-one detector (LOD), a well-established combinatorial logic module in digital circuit design. If the first 1 is found at position $k$ (counting from the most significant bit, $k = 1, 2, \ldots, 10$), the RMPA loop runs $k$ times and the corresponding threshold for the emitted spike is $\theta_k = (-1)^{S} \times 2^{E-15-k}.$ That is, the position $k$ directly encodes the scaling factor $2^{-k}$ relative to the initial threshold, without any explicit computation. After the spike is emitted, the bit at position $k$ is cleared, and the LOD is applied to the updated mantissa in the next cycle to find the next 1. This process repeats until all cycles are exhausted or the mantissa becomes zero. Unlike the traditional ANN-to-SNN approaches that require the full subtract-compare-conditional updates, this method simplifies each cycle to LOD lookups followed by a bit-clear operation, significantly reducing computational cost.


\begin{figure}[ht]
    \centering
    \includegraphics[width=1.0\linewidth]{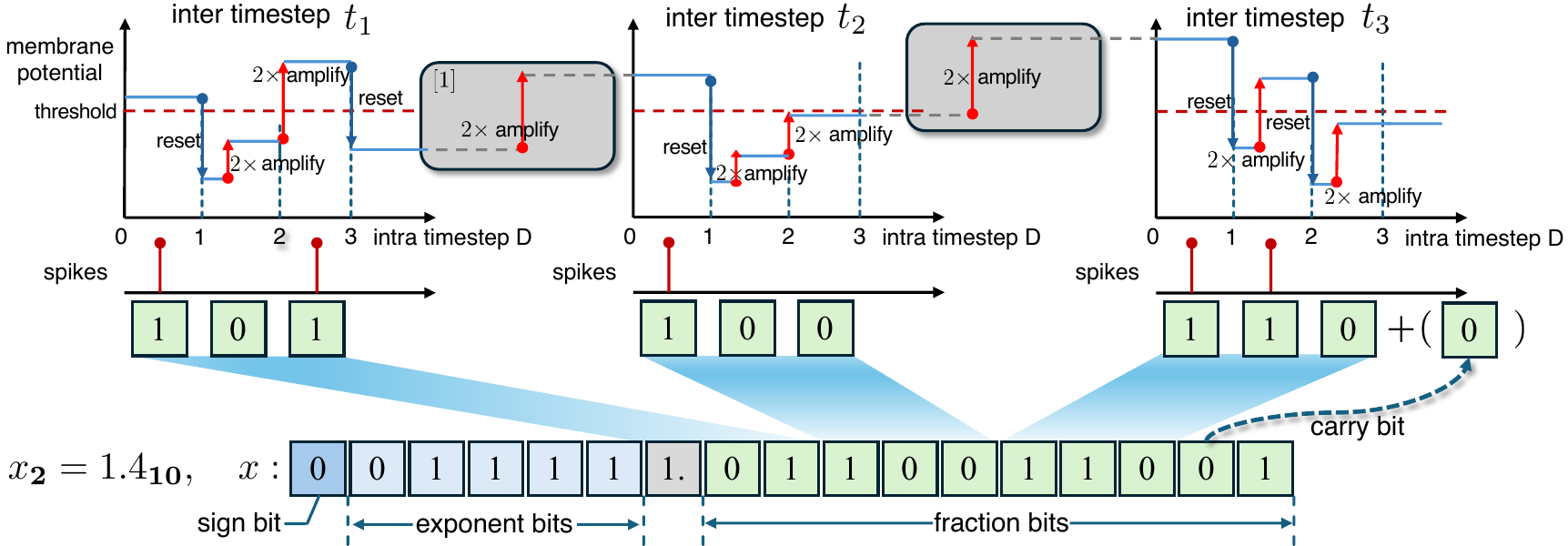}
    \caption{Parsing the FP16 representation of 1.4 into spikes with D=T=3. The first inter-timestep emits the implicit 1 concatenated with the leading 2 mantissa bits at threshold $\theta = (-1)^{S} \times 2^{E-15-(D-1)}$; the second uses the LOD to locate the first 1 in the remaining 8 bits, reads D bits from that position as the spike value, and shifts the threshold accordingly. $^{[1]}$ The number of amplifications is not strictly one. It equals the bit-offset of the next leading 1 from the end of the current $T$-bit group.}
    \label{fig:spikes2fp16}
\end{figure}

We present an example of parsing the FP16 data of 1.4 in Figure~\ref{fig:spikes2fp16}. We refer to $D$ as the intra-timestep spike dimension, which controls how many bits are emitted within a single timestep, and $T$ as the number of inter-timesteps, which controls how many successive timesteps are used to encode a given value. In the basic case ($D=1$), each timestep emits a 1-bit spike indicating whether the threshold is met. This paradigm further supports multi-bit spike emission by reading $D$ consecutive bits at a time, rather than a single bit. When $D > 1$, each spike is a $D$-bit unsigned integer ranging from 0 to $2^D - 1$, representing 0 to $2^D - 1$ times the corresponding threshold. For example, when $D=2$, 2 bits are read per timestep, encoding spike values in $\{0, 1, 2, 3\}$; when $D=3$, 3 bits are read, encoding values in $\{0, 1, \ldots, 7\}$. In the last timestep, an additional bit (i.e., the (D+1)-th bit) is read beyond the D-bit spike. This extra bit is then added to the D-bit spike value, treated as an unsigned integer, to perform a carry operation, which is equivalent to a round-half-up operation, and further improves encoding accuracy without introducing a separate rounding stage.

At the hardware level, this intra-timestep expansion only requires widening the read window from 1 bit to $D$ bits; the LOD-based position finding and threshold determination remain unchanged. This multi-bit emission significantly increases the information density per timestep and is difficult to achieve in traditional ANN-to-SNN conversion algorithms, where each timestep is inherently limited to a binary fire-or-not decision.

A key advantage of this binary-parsing approach is that when $T \leq 2$, the two timesteps are data-independent: the first reads the implicit 1 and leading mantissa bits, while the second operates on the remaining bits through a LOD. Since the second timestep depends only on a predetermined bit range, both can be executed in parallel, effectively halving the latency.

\subsubsection{Efficient Multiplication Under Spike Traces}
The spike generation mechanism described above decomposes an FP16 value into a sequence of weighted spikes over $T$ inter-timesteps. We now show how this decomposition naturally induces a hardware-efficient multiplication scheme: instead of performing a full $11 \times 11$-bit mantissa multiplication, the product $x \times w$ is computed incrementally through spike-driven conditional shift-add operations across timesteps.

\paragraph{Sign and Exponent.}
Given two FP16 operands $x$ and $w$, the sign and exponent of the product are computed in the standard manner. The output sign is obtained via a single XOR gate: $S_o = S_x \oplus S_w$, 
and the output exponent is $E_o = E_x + E_w - \text{bias}$, 
where $\text{bias} = 01111_2 = 15_{10}$. This requires only a 5-bit adder and a 6-bit subtractor, identical to a conventional FP16 multiplier. The key innovation lies entirely in how the mantissa product is computed, as described below.

\begin{figure}[ht]
    \centering
    \includegraphics[width=1.0\linewidth]{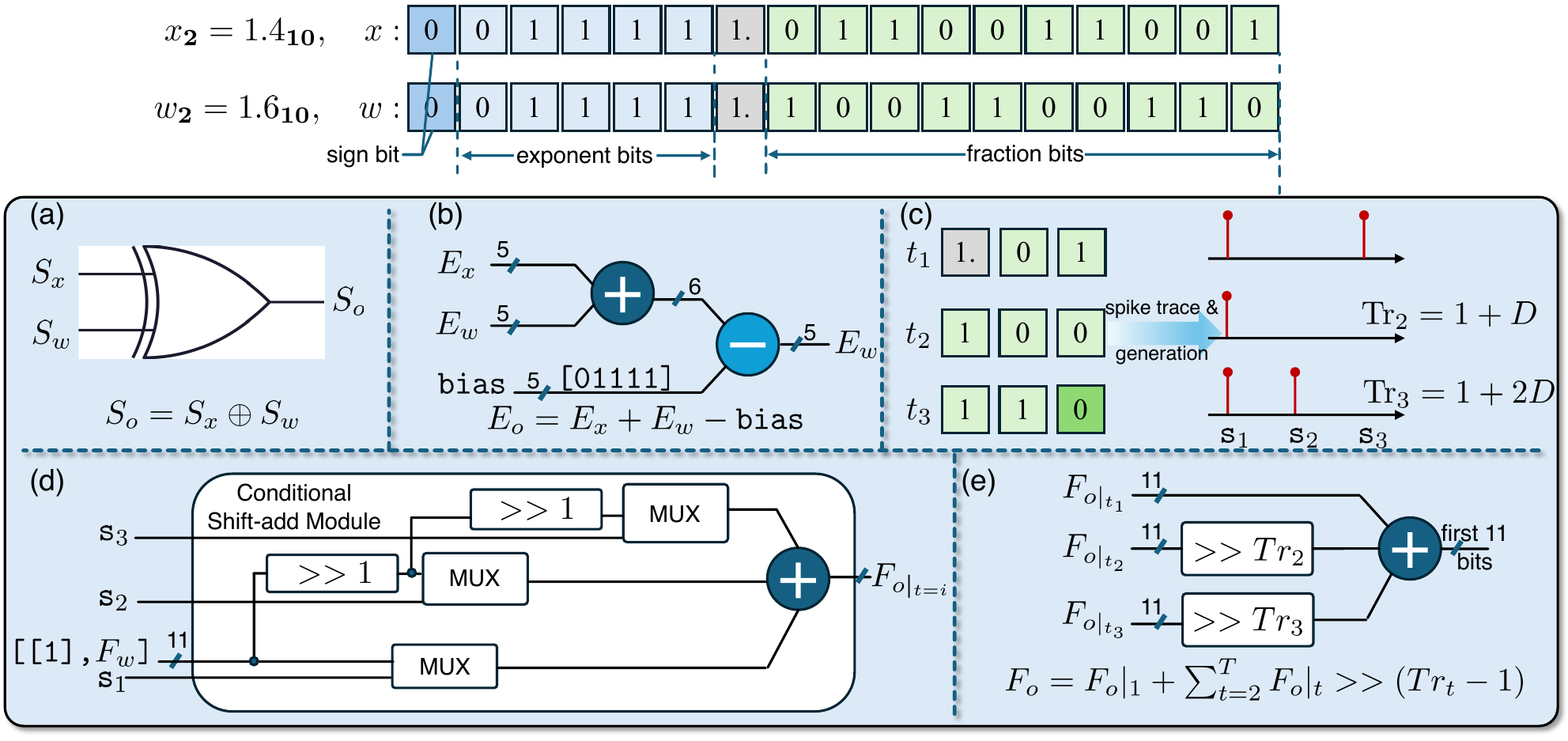}
    \caption{Spike-driven FP16 multiplication. Sign, exponent, and mantissa of $x$ and $w$ are processed in parallel: \textbf{(a)} $S_o = S_x \oplus S_w$ via XOR; \textbf{(b)} $E_o = E_x + E_w - \text{bias}$ via a 5-bit adder and 6-bit subtractor; \textbf{(c)} the mantissa $F_x$ is decomposed into $T$ spikes $\{s_t\}$ with traces $\mathrm{Tr}_t = 1+(t-1)D$; \textbf{(d)} each partial product $F_{o}|_t = s_t \times F_w$ is computed by a Conditional Shift-add Module of MUXes and an adder tree, replacing the $11\times11$ multiplier array; \textbf{(e)} partial products are trace-aligned via right shifts and summed to form $F_o$, truncated to 11 bits.}
    \label{fig:calc}
\end{figure}

\paragraph{Mantissa Decomposition via Spike Traces.}
Let $F_x = 1.M_{x,1} M_{x,2} \cdots M_{x,10}$ denote the full mantissa of $x$ (including the implicit leading 1), and similarly $F_w$ for $w$. The spike generation process of the previous subsection decomposes $F_x$ into $T$ spike values $\{s_t\}_{t=1}^{T}$, each being a $D$-bit unsigned integer, with an associated trace value $\mathrm{Tr}_t$ that records the bit-position offset of the $t$-th spike relative to the binary point. Specifically, the first spike $s_1$ is composed of the implicit 1 together with the first $D-1$ mantissa bits, and the corresponding trace is $\mathrm{Tr}_1 = 1$. For subsequent timesteps $t \geq 2$, the LOD identifies the starting position of the next group, and $D$ consecutive bits are read as $s_t$. The trace value satisfies:
$
\mathrm{Tr}_t = 1 + (t-1) \times D,
$
when the mantissa bits are densely packed (i.e., no intervening zeros between groups). Under this decomposition, the mantissa can be reconstructed as $F_x = \sum_{t=1}^{T} s_t \times 2^{-(\mathrm{Tr}_t - 1)}.$ 
Substituting into the mantissa product yields:
\begin{equation}\label{eq:mantissa_product}
F_o = F_x \times F_w = \sum_{t=1}^{T} \underbrace{(s_t \times F_w)}_{\triangleq\; F_{o}|_t} \times 2^{-(\mathrm{Tr}_t - 1)}.
\end{equation}
This factorization reveals that the $11 \times 11$-bit mantissa multiplication is decomposed into $T$ smaller multiplications of a $D$-bit spike $s_t$ by the 11-bit mantissa $F_w$, followed by trace-aligned accumulation. Each per-timestep partial product $F_{o}|_t = s_t \times F_w$ is a $D$-bit $\times$ 11-bit multiplication, which can be efficiently realized by a Conditional Shift-add Module as illustrated in Figure~\ref{fig:spikes2fp16}(d). Let $s_t = \sum_{j=1}^{D} b_j \times 2^{D-j}$, where $b_j \in \{0,1\}$ are the individual bits of the $D$-bit spike. Then:
\begin{equation}\label{eq:shift_add}
F_{o}|_t = s_t \times F_w = \sum_{j=1}^{D} b_j \times (F_w \gg (j-1)),
\end{equation}
where $\gg$ denotes logical right shift. Each term $b_j \times (F_w \gg (j-1))$ is implemented by a multiplexer (MUX) controlled by $b_j$: when $b_j = 1$, the shifted value $F_w \gg (j-1)$ is selected; when $b_j = 0$, zero is selected. The $D$ selected values are then summed by an adder tree. This conditional shift-add structure replaces the combinatorial multiplier array with $D$ MUXes and a $(D-1)$-stage adder tree, significantly reducing the hardware area. Note that when $\text{spike} = 111$, the multiplication can be computed more efficiently by rewriting it as $1000 - 0001$, requiring only a single subtraction. A detailed derivation is provided in Appendix~\ref{sec:111}.
 
\paragraph{Trace-aligned Temporal Accumulation.}
As shown in Figure~\ref{fig:calc}(e), the final mantissa is the trace-aligned sum of the $T$ partial products $F_{o}|_t$: 
\begin{equation}\label{eq:temporal_accum}
F_o = F_{o}|_1 + \sum_{t=2}^{T} F_{o}|_t \gg (\mathrm{Tr}_t - 1).
\end{equation}
The right shift $\gg (\mathrm{Tr}_t - 1)$ aligns the $t$-th partial product to the correct bit position relative to the first partial product, and the final sum is truncated to the first 11 bits to produce the output mantissa. This temporal accumulation directly mirrors the dynamics of synaptic integration in SNNs: each incoming spike $s_t$ triggers a weighted response $F_{o}|_t$, and the trace $\mathrm{Tr}_t$ governs the temporal decay (realized here as a right shift) before the responses are accumulated at the output neuron.

As shown in appendix~\ref{sec:energy},  under the setting T = 2, D = 3, the Spike Generation via Direct FP16 Binary Parsing and the Spike-driven Multiplication help make the process cost only 13.4\% of the energy consumption of FP32 multiplication and 40.6\% of the bit-transmission cost between layers of FP32.
\section{Experiments}


In this section, we evaluate the proposed PRIME framework on both computer vision (CV) and natural language processing (NLP) tasks, comparing conversion accuracy against existing approaches. We further examine how different parameter settings affect performance through ablation studies. All experiments are performed on a single NVIDIA A100 GPU with PyTorch 2.10.0 and CUDA 12.8.

\subsection{Task Evaluation on Computer Vision}

For CV tasks, we validate the effectiveness of PRIME on representative CNN and Transformer architectures, including VGG, ResNet, ViT, and EVA, evaluated on the ImageNet-1K~\cite{Imagenet} and CIFAR~\cite{cifar} (provided in Appendix~\ref{overall_cifars}) datasets. We follow the experimental settings of~\cite{ECMT}.

The experimental results on ImageNet-1K are presented in Table~\ref{tab:comparison}. PRIME consistently achieves lower conversion error across both CNN and Transformer architectures. Compared with the most representative baseline DCGS~\cite{dcgs}, PRIME requires only two timesteps to attain lower accuracy degradation on VGG and ResNet ($0.04\%$ vs. $0.08\%$, $0.00\%$ vs. $0.38\%$), and demonstrates even more pronounced advantages on the ViT and EVA architectures ($0.00\%$ vs. $0.39\%$, $0.00\%$ vs. $0.33\%$). In summary, PRIME is the closest to lossless ANN-to-SNN conversion among existing frameworks, while further reducing the inter-layer bit-width during hardware deployment ($13$ bits vs. $16$ bits).

\begin{table*}[htbp]
  \centering
  \caption{Performance comparison of different methods on the ImageNet-1K dataset. ``Bits'' denotes the total number of bits transmitted across all timesteps (i.e., the cumulative bit-width of the spike over $T$ steps), reflecting the inter-layer data transmission cost during inference.}
  \label{tab:comparison}
  \resizebox{\textwidth}{!}{
  \begin{tabular}{llccccccc}
    \toprule
    Method & Arch. & Param.(M) & ANN Acc.(\%) & $T$ & Bits & SNN Acc.(\%) \\
    \midrule
    TS\cite{TS}
      & VGG-16         & 138.0    & 72.40 & 64           & 64          & 70.97~(-1.43) \\
    SNM\cite{SNM}
      & VGG-16         & 138.0    & 73.18 & 64           & 128         & 71.50~(-1.68) \\
    \cmidrule(lr){2-7}
    \multirow{2}{*}{QCFS\cite{QCFS}}
      & ResNet-34      & 21.8   & 74.32 & 64           & 64          & 72.35~(-1.97) \\
      & VGG-16         & 138.0    & 74.29 & 64           & 64          & 72.85~(-1.44) \\
    \cmidrule(lr){2-7}
    \multirow{2}{*}{SRP\cite{SRP}}
      & ResNet-34      & 21.8   & 74.32 & 64        & 64       & 68.61~(-5.71) \\
      & VGG-16         & 138.0    & 74.29 & 64        & 64       & 69.43~(-4.86) \\
    \cmidrule(lr){2-7}
    MST\cite{MST}
      & Swin-T         & 28.5   & 80.51 & 512     & 512    & 78.51~(-2.00) \\
    STA\cite{STA}
      & ViT-B/32       & 86.0     & 83.60 & 256      & 256     & 82.79~(-0.81) \\
    \cmidrule(lr){2-7}
    \multirow{3}{*}{SpikeZIP-TF\cite{spikeziptf}}
      & SViT-S-32 & 22.1  & 81.59 & 64           & 128         & 81.45~(-0.14) \\
      & SViT-B-32 & 86.6  & 82.83 & 64           & 128         & 82.71~(-0.12) \\
      & SViT-L-32 & 304.3 & 83.86 & 64           & 128         & 83.82~(-0.04) \\
    \cmidrule(lr){2-7}
    \multirow{2}{*}{ECMT\cite{ECMT}}
      & ViT-S/16       & 22.0     & 78.04 &10        & 40      & 77.07~(-0.97) \\
      & EVA-G          & 1074.0   & 89.62 &8         & 32      & 89.40~(-0.22) \\
    \cmidrule(lr){2-7}
    \makecell[l]{Adaptive fission\cite{adaptive}\\ + QCFS\cite{QCFS}}
      & ResNet-34      & 21.8   & 74.32 & 32           & 64          & 69.37~(-4.95) \\
    \cmidrule(lr){2-7}
    \multirow{4}{*}{DCGS\cite{dcgs}}
      & VGG-4/1        & 138.0    & 73.25 &  8         &  24      &  73.17~(-0.08) \\
      & ResNet34-4/1   & 21.8   & 76.42 & 8         &  24      &  76.04~(-0.38) \\
      & ViT-L-8/4      & 304.3  & 85.84 &  4         &  16       &  85.45~(-0.39) \\
      & EVA02-L-8/4    & 305.1  & 90.05 &  4         &  16       & 89.72~(-0.33) \\
    \midrule
    \multirow{10}{*}{\textbf{PRIME~(ours)}}
      & VGG-16         & 138.0    & 71.12 & \textbf{2} $(D{=}3)$  & \textbf{13}          & \textbf{71.08~(-0.04)} \\
      & ResNet-34      & 21.8   & 76.09 & \textbf{2} $(D{=}3)$  & \textbf{13}          & \textbf{76.13~(-0.00)} \\
      & ViT-S          & 22.1   & 78.04 & \textbf{2} $(D{=}3)$  & \textbf{13}          & \textbf{78.06~(-0.00)} \\
      & ViT-B          & 86.6   & 82.91 & \textbf{2} $(D{=}3)$  & \textbf{13}          & \textbf{82.87~(-0.04)} \\
      & ViT-L          & 304.3  & 84.88 & \textbf{2} $(D{=}3)$  & \textbf{13}          & \textbf{84.89~(-0.00)} \\
      & EVA02-T        & 5.8    & 80.62 & \textbf{2} $(D{=}3)$  & \textbf{13}          & \textbf{80.63~(-0.00)} \\
      & EVA02-S        & 22.1   & 85.73 & \textbf{2} $(D{=}3)$  & \textbf{13}          & \textbf{85.70~(-0.03)} \\
      & EVA02-B        & 87.1   & 88.26 & \textbf{2} $(D{=}3)$  & \textbf{13}          & \textbf{88.27~(-0.00)} \\
      & EVA02-L        & 305.1  & 89.63 & \textbf{2} $(D{=}3)$  & \textbf{13}          & \textbf{89.63~(-0.00)} \\
      & EVA02-G        & 1013.0   & 89.62 & \textbf{2} $(D{=}3)$  & \textbf{13}          & \textbf{89.58~(-0.04)} \\
    \bottomrule
  \end{tabular}
  }
\end{table*}

\subsection{Task Evaluation on Natural Language}

For NLP tasks, we apply our method to the Qwen and Llama series models~\cite{qwen3,llama2}, and perform evaluations on the main benchmarks, including MMLU, PIQA, ARC-C, ARC-E, Winogrande, and Hellaswag~\cite{mmlu,piqa,wiqa,arc}. The evaluation process is implemented based on the work of~\cite{eval-harness}. We evaluate our method on common sense reasoning tasks using the LLAMA2-7b and LLAMA2-13b models with ultra-low latency (\(T=2\)). Following common practices in SNN conversion, we adopt the same architectural configurations without modifications. The results are presented in Table~\ref{tab:commonsense}.
To the best of our knowledge, this is the first conversion framework that consistently attains ‑0.3\% accuracy degradation on common sense reasoning benchmarks with only 2 timesteps.
More results under different settings and datasets are shown in appendix~\ref{overall_mmlu} and~\ref{overall_llama}.


\begin{table*}
  \centering
  \caption{Performance Comparison on Common Sense Reasoning Tasks (\%). ``Bits'' denotes the total bits transmitted across all timesteps, reflecting the data transmission cost during inference.}
  \label{tab:commonsense}
  \resizebox{\textwidth}{!}{
  \begin{tabular}{llcccccccccc}
    \toprule
    Method & Model & Spike & $T$ & Bits & PIQA & ARC-c\_n & ARC-e\_n & Winog. & HellS.\_n \\
    \midrule

    \multirow{2}{*}{LLAMA}
      & LLAMA2-7b  & $\times$ & N/A          & 16 & 78.45 & 45.73 & 74.87 & 69.38 & 76.28 \\
      & LLAMA2-13b & $\times$ & N/A          & 16 & 80.52 & 48.89 & 76.56 & 72.22 & 79.71 \\

    \multirow{2}{*}{SpikeLLM\cite{spikellm}}
      & LLAMA2-7b  & $\checkmark$ & 2        & 32 & 65.67 & 28.41 & 41.88 & 52.80 & 49.87 \\
      & LLAMA2-13b & $\checkmark$ & 2        & 32 & 67.68 & 30.38 & 44.53 & 57.22 & 58.02 \\

    FAS\cite{fas}
      & OPT-7B     & $\checkmark$ & 16       & 32 & 73.23 & ---   & 64.73 & 60.38 & ---   \\

    \multirow{3}{*}{LAS\cite{las}}
      & OPT-7B     & $\checkmark$ & 16       & 64 & 76.22 & ---   & 65.95 & 65.75 & ---   \\
      & OPT-13B    & $\checkmark$ & 16       & 64 & 76.28 & ---   & 67.38 & 65.51 & ---   \\
      & OPT-66B    & $\checkmark$ & 16       & 64 & 78.67 & ---   & 71.25 & 68.27 & ---   \\

    \cmidrule{1-10}

    \multirow{2}{*}{PRIME~(ours)}
      & LLAMA2-7b  & $\checkmark$ & \textbf{2 $(D{=}3)$} & \textbf{13} & \textbf{78.67} & \textbf{45.65} & \textbf{74.75} & \textbf{69.06} & \textbf{76.18} \\
      & LLAMA2-13b & $\checkmark$ & \textbf{2 $(D{=}3)$} & \textbf{13} & \textbf{80.58} & \textbf{49.15} & \textbf{76.47} & \textbf{72.14} & \textbf{79.69} \\

    \bottomrule
  \end{tabular}
  }
\end{table*}
\subsection{Ablation Study}

By default, we adopt the full FP16 exponent bit-width (5 bits) to set the threshold range, which preserves conversion accuracy. To evaluate the trade-off between efficiency and accuracy, we reduce the exponent bit-width to 4 and report the results in Table~\ref{tab:ablation2}. Across all $(T, D)$ configurations, the accuracy drop is marginal, while the number of active thresholds and the inter-layer bit-transmission cost is substantially reduced. This confirms that the proposed threshold-range optimization offers a practical knob for resource-constrained deployment with negligible performance sacrifice.

\begin{table*}[htbp]
  \centering
  \scriptsize  
  \caption{Ablation study on lightweight threshold-range optimization via exponent-bit 
  width reduction (\texttt{width=4}) on ImageNet-1K, evaluated across different 
  $(T, D)$ settings. \texttt{Default} refers to the full fp16 exponent-bit width 
  , with spike conversion applied to $k$ and $v$ only in Transformer 
  blocks.}
  \label{tab:ablation2}
  \resizebox{\textwidth}{!}{
  \begin{tabular}{cccccccccc}
    \toprule
    \multirow{2}{*}{Model} & \multirow{2}{*}{ANN} & \multirow{2}{*}{D} 
    & \multicolumn{3}{c}{Default} 
    & \multicolumn{3}{c}{Exponent-bit width $= 4$} \\
    \cmidrule(lr){4-6} \cmidrule(lr){7-9}
    & & & $T=1$ & $T=2$ & $T=3$ & $T=1$ & $T=2$ & $T=3$ \\
    \midrule
    \multirow{3}{*}{ViT-S (patch16\_224)} 
    & \multirow{3}{*}{78.04} 
    & 1 & 48.20 & 76.27 & 77.80 & 55.26 & 76.29 & 77.47 \\
    & & 2 & 73.78 & 77.94 & 78.03 & 73.94 & 77.66 & 77.65 \\
    & & 3 & 77.01 & 78.06 & 78.03 & 76.92 & 77.66 & 77.65 \\
    \midrule
    \multirow{3}{*}{ViT-B (patch16\_224)}
    & \multirow{3}{*}{82.91} 
    & 1 & 74.00 & 82.21 & 82.74 & 75.80 & 81.30 & 81.77 \\
    & & 2 & 80.89 & 82.77 & 82.89 & 80.30 & 81.74 & 81.88 \\
    & & 3 & 82.41 & 82.87 & 82.91 & 81.38 & 81.87 & 81.87 \\
    \midrule
    \multirow{3}{*}{ViT-L (patch16\_224)} 
    & \multirow{3}{*}{84.88} 
    & 1 & 80.89 & 84.46 & 84.79 & 81.14 & 83.81 & 84.31 \\
    & & 2 & 84.03 & 84.80 & 84.89 & 83.25 & 84.31 & 84.32 \\
    & & 3 & 84.45 & 84.89 & 84.88 & 84.01 & 84.30 & 84.30 \\
    \bottomrule
  \end{tabular}
  }
\end{table*}
\section{Conclusion and Limitations}
In this paper, we propose PRIME, a unified ANN-to-SNN conversion framework that jointly addresses information redundancy in time-step extension and computational redundancy in auxiliary neuronal dynamics. At the encoding level, integration-coding-based MTNs leverage the RMPA mechanism to actively amplify residual membrane potential at each iteration, achieving an exponentially tighter error bound of $\theta/2^{T \cdot D+1}$ versus the linear bound $\theta/2^{T}$ of rate coding, enabling near-lossless conversion at $T \leq 2$. At the inference level, spike thresholds and values are read directly from the floating-point binary representation, eliminating membrane potential accumulation and multi-threshold comparison; the resulting spike-driven multiplication requires at most a single adder per cycle when $D \leq 3$. Under $T = 2$, $D = 3$, energy consumption reaches 13.4\% of FP32 multiplication and bit-transmission cost 40.6\% of FP32. These figures rely on analytical estimation; however, all underlying primitives (bit-shift, integer addition, and leading-one detection) are natively supported by standard digital logic, making physical implementation a natural next step.
\bibliographystyle{unsrt}
\bibliography{reference}
\clearpage

\appendix

\section{Proof of Proposition~\ref{prop:rmpa-mtn-error-bound}\label{sec:error}}
\begin{proof}
The conversion error in ANN-to-SNN algorithms arises primarily from residual membrane potential below the threshold that fails to generate corresponding spikes.

\subsection*{Conventional Rate Coding}

In conventional ANN-to-SNN conversion, the neuron accumulates the input signal $\mathbf{I}$ at each timestep:
\[
\mathbf{m} \leftarrow \mathbf{m} + \mathbf{I}.
\]
The membrane potential is then compared against the threshold $\theta$. If $\mathbf{m} > \theta$, a spike is emitted and the membrane potential is updated as:
\[
\mathbf{m} \leftarrow \mathbf{m} - \theta.
\]
After $T$ accumulation steps:
\[
T \cdot \mathbf{I} = \theta \cdot N_{\text{spike}} + \mathbf{m}_{\text{rest}},
\]
where $N_{\text{spike}}$ denotes the total number of emitted spikes. The ideal encoded value is $\mathbf{I}$, whereas the actual decoded value is:
\[
\frac{\theta \cdot N_{\text{spike}}}{T} = \mathbf{I} - \frac{\mathbf{m}_{\text{rest}}}{T}.
\]
The encoding error over $T$ steps is therefore:
\[
\varepsilon = \frac{\mathbf{m}_{\text{rest}}}{T}, \quad \mathbf{m}_{\text{rest}} < \theta \implies \varepsilon < \frac{\theta}{T}.
\]
With optimized initial membrane potential initialization, $\mathbf{m}_{\text{rest}} < \theta/2$ can be guaranteed~\cite{hao2023bridging}, yielding a tighter bound:
\[
\varepsilon < \frac{\theta}{2T}.
\]

\subsection*{Proposed Integration Coding}

The method proposed in this paper adopts integration coding, where the input signal $\mathbf{I}$ is applied only at the first timestep:
\[
\mathbf{m} = \mathbf{I}.
\]
In subsequent steps, the residual membrane potential is amplified without further input injection. Let $n_i$ denote the amplification index at which the $i$-th spike is emitted; the corresponding spike amplitude is $\theta / 2^{n_i}$. After $n$ amplification operations, the input is decomposed as:
\[
\mathbf{I} = \sum_{i} \frac{\theta}{2^{n_i}} + \frac{\mathbf{m}_{\text{rest}}}{2^{n}},
\]
where the second term represents the residual encoding error. The encoding error is thus:
\[
\varepsilon = \frac{\mathbf{m}_{\text{rest}}}{2^{n}}, \quad \mathbf{m}_{\text{rest}} < \theta \implies \varepsilon < \frac{\theta}{2^{n}}.
\]
When the carry mechanism is applied, the error is further reduced by half:
\[
\varepsilon < \frac{\theta}{2^{n+1}}.
\]
Here, $n$ denotes the total number of amplification operations. When $T = 1$, we have $n = D$. When $T > 1$, additional amplification operations are required between timesteps to satisfy the condition $\mathbf{m}_{\text{rest}} > \theta$ prior to spike emission, yielding:
\[
n \geq T \cdot D,
\]
and the error is bounded by:
\[
\varepsilon < \frac{\theta}{2^{T \cdot D + 1}}.
\]
\end{proof}



\section{Energy Consumption Analysis\label{sec:energy}}

The energy savings achieved by ANN-to-SNN conversion stem from two primary sources: (1) replacing floating-point multiplication with lightweight addition operations, and (2) reducing bit-transmission cost between layers through the proposed spike encoding scheme. This section provides a detailed quantitative analysis of both aspects.

\subsection{Computational Energy Consumption}

Based on the proposed spike acquisition method, all multiplication operations $w \times x$ are fully decomposed into integer-level additions and bit-shift operations. For reference, prior work has established that FP32 multiplication consumes 3.7\,pJ and FP16 multiplication consumes 1.1\,pJ~\cite{energy}, as summarized in Table~\ref{tab:energy_operation}.

\begin{table}[htbp]
    \centering
    \caption{Energy Consumption of Common Hardware Operations}
    \label{tab:energy_operation}
    \begin{tabular}{ccc ccc}
        \toprule
        \multicolumn{2}{c}{\textbf{Integer Operation}}
        & \multicolumn{2}{c}{\textbf{Floating-Point Operation}}
        & \multicolumn{2}{c}{\textbf{Memory Access (64-bit)}} \\
        \cmidrule(lr){1-2} \cmidrule(lr){3-4} \cmidrule(lr){5-6}
        Operation   & Energy   & Operation    & Energy   & Memory       & Energy        \\
        \midrule
        8-bit Add   & 0.03\,pJ & 16-bit FAdd  & 0.4\,pJ  & 8\,KB Cache  & 10\,pJ        \\
        32-bit Add  & 0.1\,pJ  & 32-bit FAdd  & 0.9\,pJ  & 32\,KB Cache & 20\,pJ        \\
        8-bit Mult  & 0.2\,pJ  & 16-bit FMult & 1.1\,pJ  & 1\,MB Cache  & 100\,pJ       \\
        32-bit Mult & 3.1\,pJ  & 32-bit FMult & 3.7\,pJ  & DRAM         & 1.3--2.6\,nJ  \\
        \bottomrule
    \end{tabular}
\end{table}

Taking $T=2$, $D=3$ as a representative configuration (illustrated in Fig.~\ref{fig:calc} with $x=1.4$, $w=1.6$), the computation of $x \times w$ proceeds as follows.

\textbf{Step 1: Spike extraction from $x$.}
\begin{itemize}
    \item \textit{Cycle 1}: The spike is formed by the implicit leading 1 of the mantissa concatenated with the first $D-1=2$ fractional bits.
    \item \textit{Cycle 2}: The position $\mathrm{Tr}_2 $ of the first set bit among the remaining 8 mantissa bits is identified (here $\mathrm{Tr}_2=1 $), yielding a raw spike of \texttt{100}. A carry operation is then applied: the bit immediately following \texttt{100} is 1, so the spike is corrected to $\texttt{100}+1=\texttt{101}$.
    \item \textit{Remark}: Since the leading bit of each spike is always 1 by construction, it need not be transmitted explicitly; only the trailing $D-1$ bits are communicated, further reducing  bit-transmission cost between layers.
\end{itemize}

\textbf{Step 2: Exponent and sign processing.}
The sign bit of the result is obtained by XOR-ing the sign bits of $w$ and $x$. The exponent of the result is computed by summing the exponent fields of $w$ and $x$ and subtracting the bias of 15. The fractional part of $w$ is then shifted and accumulated in parallel according to the spike value of each cycle:
\begin{itemize}
    \item \texttt{spike = 101}: add the mantissa of $w$ and its right-shift by 2 bits.
    \item \texttt{spike = 100}: retain the mantissa of $w$ unchanged.
    \item \texttt{spike = 110}: add the mantissa of $w$ and its right-shift by 1 bit.
    \item \texttt{spike = 111}: compute the difference between the left-shifted mantissa and its right-shift by 2 bits (realizable via two's complement addition).
\end{itemize}

\textbf{Step 3: Cross-cycle accumulation.}
The fractional result from Cycle 1 is added to that of Cycle 2 right-shifted by $(\mathrm{Tr}_2 - 1)$ bits, accounting for any threshold scaling introduced by the carry operation. The final product $w \times x$ is assembled by concatenating the sign bit, the exponent, and the leading 11 bits of the accumulated mantissa.

\subsubsection*{Energy Estimation Basis}

Following established hardware scaling rules, energy consumption scales sub-linearly with operand width: doubling the bit-width typically increases energy by less than $2\times$~\cite{energy_scale}. On this basis, INT16 addition is estimated at 0.05--0.06\,pJ (conservatively taken as 0.06\,pJ), and sign-bit XOR is upper-bounded by INT8 addition at 0.03\,pJ.

\paragraph{First-Cycle Energy}

\begin{enumerate}[label=(\arabic*)]
    \item \textbf{Sign-bit XOR}: $0.03\,\text{pJ}$.
    \item \textbf{Exponent summation}: summing exponent bits and subtracting the bias consumes $0.06\,\text{pJ}$.
    \item \textbf{Spike--mantissa multiplication}:
    \begin{enumerate}[label=\textit{D=\arabic*}:, leftmargin=*]
        \item Retain mantissa directly; $0\,\text{pJ}$.
        \item With probability $1/2$ (\texttt{spike=10}): retain directly ($0\,\text{pJ}$); with probability $1/2$ (\texttt{spike=11}): one INT16 addition. Average: $(0+0.06)/2 = 0.03\,\text{pJ}$. Total with sign/exponent: $0.045\,\text{pJ}$.
        \item With probability $1/4$ (\texttt{spike=100}): $0\,\text{pJ}$; with probability $1/2$ (\texttt{spike=101,110}): one INT14 addition ($\approx0.06\,\text{pJ}$); with probability $1/4$ (\texttt{spike=111}): complement-based subtraction (estimated at $1.5\times$ INT addition). Average: $0.075\,\text{pJ}$.
    \end{enumerate}
    \item \textbf{Rounding}: one rounding operation on the mantissa at the end of each multiplication, $0.03\,\text{pJ}$, overlapped with Cycle 1.
\end{enumerate}

\paragraph{Subsequent-Cycle Energy ($T \geq 2$)}

\begin{enumerate}[label=(\arabic*)]
    \item \textbf{Leading-one detection}: locating the first set bit and recording offset $p$ is negligible; upper-bounded by INT8 addition at $0.03\,\text{pJ}$.
    \item \textbf{Spike--mantissa multiplication}: identical rules as Cycle 1; average energy is $0\,\text{pJ}$, $0.045\,\text{pJ}$, and $0.075\,\text{pJ}$ for $D=1,2,3$, respectively.
    \item \textbf{Cross-cycle accumulation}: right-shift by $(D+Tr_2-1)$ bits followed by addition. The maximum operand widths are INT24 ($D=2$) and INT25 ($D=3$), both truncated to INT16 addition: $0.03+0.06=0.09\,\text{pJ}$.
\end{enumerate}

The energy requirements for all subsequent cycles ($T>2$) are identical to those of Cycle 2. The total energy consumption in representative configurations $(T, D)$ is reported in Table~\ref{tab:power_consumption}.

\begin{table}[htbp]
    \centering
    \caption{Total Energy Consumption and Energy Ratio Relative to FP32 for Different $(T,\,D)$ Configurations (Unit: pJ)}
    \label{tab:power_consumption}
    \resizebox{0.5\textwidth}{!}{%
    \begin{tabular}{lccc}
        \toprule
        \textbf{Energy (pJ)} & $T{=}1$ & $T{=}2$ & $T{=}3$ \\
        \midrule
        $D{=}1$ & 0.090 & 0.165 & 0.195 \\
        $D{=}2$ & 0.210 & 0.330 & 0.390 \\
        $D{=}3$ & 0.330 & 0.495 & 0.585 \\
        \midrule
        FP16 (ANN) & \multicolumn{3}{c}{1.100} \\
        FP32 (ANN) & \multicolumn{3}{c}{3.700} \\
        \bottomrule
    \end{tabular}}
    \vspace{0.8em}
    \resizebox{0.5\textwidth}{!}{%
    \begin{tabular}{lccc}
        \toprule
        \makecell{\textbf{Energy Ratio} \\ \textbf{(SNN/FP32)}} & $T{=}1$ & $T{=}2$ & $T{=}3$ \\
        \midrule
        $D{=}1$ & 0.024 & 0.045 & 0.053 \\
        $D{=}2$ & 0.057 & 0.089 & 0.105 \\
        $D{=}3$ & 0.089 & 0.134 & 0.158 \\
        \bottomrule
    \end{tabular}}
\end{table}

\subsection{ bit-transmission cost between layers Evaluation}
 In traditional schemes, the  bit-transmission cost between layers depends on the data type: if the FP32 format is adopted, 32-bit data needs to be transmitted for each multiplication calculation; if the FP16 or BF16 format is adopted, 16-bit data needs to be transmitted for each multiplication calculation.
 The proposed  bit-transmission cost between layers consists of three parts: 1 bit sign information, 5 bit threshold information, $(D-1)$ bit spike information transmitted in the first cycle (the first bit defaults to 1 and can be omitted), and $(D-1+3)$ bit spike information transmitted in subsequent cycles (where the additional 3 bits are used to represent the threshold offset $\mathrm{Tr}_t$, with an upper limit of 8 bits: 000 for 1 and 111 for 8, and offsets exceeding 8 bits are directly ignored). The detailed transmission volume in different configurations is summarized in Table~\ref{tab:data_transmission}.
  \begin{table}[htbp]
    \centering
     \caption{ Bit-transmission Cost Between Layers Under Different Exponent Bit Widths and Parameter Configurations}
  \begin{tabular}{ccccc}
 
    \toprule
     \textbf{exp\_dim} & $\boldsymbol{D}$ & $\boldsymbol{T{=}1}$ & $\boldsymbol{T{=}2}$ & $\boldsymbol{T{=}3}$ \\
     \midrule
     \multirow{3}{*}{5} 
     & 1 & 6  & 9  & 15 \\
     & 2 & 7  & 11 & 18 \\
     & 3 & 8  & 13 & 21 \\
     \midrule
     \multirow{3}{*}{4} 
     & 1 & 5  & 8  & 14 \\
     & 2 & 6  & 10 & 17 \\
     & 3 & 7  & 12 & 20 \\
     \midrule
     FP32 (ANN)  & \multicolumn{4}{c}{32} \\
     FP16 or BF16 (ANN)  & \multicolumn{4}{c}{16} \\
     \bottomrule
   \end{tabular}
   
   \label{tab:data_transmission}
   \end{table}
 As can be seen in Table~\ref{tab:data_transmission}, the transmission bit width of the proposed scheme is significantly lower than that of traditional ANN schemes, thus further saving energy consumption in the data transmission process.

\section{Special Case: \texttt{spike} = \texttt{111}}
\label{sec:111}
\begin{wrapfigure}{r}{0.4\textwidth}
    \centering
    \includegraphics[width=\linewidth]{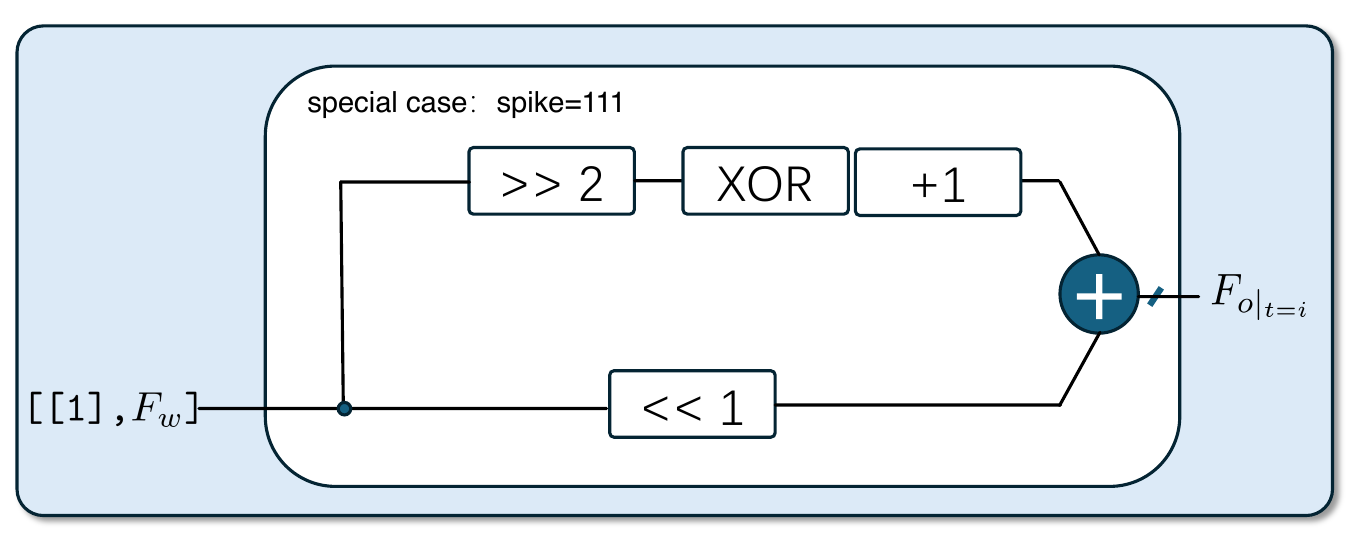}
    \caption{A special case for input spikes equals "\texttt{111}"}
    \label{fig:111}
\end{wrapfigure}
When $D = 3$ and \texttt{spike} $= \texttt{111}$, the naive shift-add formulation requires three terms:
\begin{equation}
    \begin{aligned}
        \relax[1, F_w] \cdot \texttt{111}_2 = [1, F_w] + ([1, F_w] \gg 1) + ([1, F_w] \gg 2),
    \end{aligned}
\end{equation}
which would naively require a two-stage adder tree. However, this case admits a more efficient decomposition. Observing that $\texttt{111}_2 = \texttt{1000}_2 - \texttt{0001}_2$, and noting that multiplication by $\texttt{1000}_2$ and $\texttt{0001}_2$ correspond to a left shift by 1 bit and a right shift by 2 bits respectively (relative to the implicit leading bit), we obtain:
\[
[1, F_w] \cdot \texttt{111}_2 = ([1, F_w] \ll 1) - ([1, F_w] \gg 2),
\]
which requires only a single subtraction. In hardware, subtraction is implemented via two's complement addition: negating $[1, F_w] \gg 2$ is achieved by a bitwise complement followed by adding 1 (absorbed into the carry-in of the adder), yielding:
\[
[1, F_w] \cdot \texttt{111}_2 = ([1, F_w] \ll 1) + \overline{([1, F_w] \gg 2)},
\]
where $\ll$ and $\gg$ denote left and right logical shift respectively, and $\overline{(\cdot)}$ denotes the two's complement negation. As illustrated in Fig.~\ref{fig:111}, the datapath consists of a left-shift-by-1 unit on the lower path, a right-shift-by-2 unit followed by a two's complement unit on the upper path, and a single adder combining both results.

This decomposition ensures that, for all spike patterns with $D \leq 3$, each intra timestep computation requires \textbf{at most one adder}, with no case requiring more than a single addition or subtraction operation. The hardware overhead is therefore strictly bounded regardless of the spike value, confirming the area efficiency of the proposed design.

\section{Single-Step vs.\ Multi-Step Execution Modes}
\addcontentsline{toc}{section}{Appendix A: Single-Step vs.\ Multi-Step Execution Modes}

In ANN-to-SNN conversion, two execution modes can be distinguished based on the 
temporal scheduling strategy: \textit{single-step mode} and \textit{multi-step mode}.

\paragraph{Single-Step Mode.}
In single-step mode, the input data at each time step propagates through the entire 
network sequentially, layer by layer, before the next time step begins. That is, at 
time step $t$, the spike train is fully processed across all layers---from input to 
output---prior to the commencement of computation at time step $t+1$.

\paragraph{Multi-Step Mode.}
In multi-step mode, the execution order is transposed: for each layer, computations 
across all $T$ time steps are completed before proceeding to the next layer. 
Specifically, layer $\ell$ processes inputs for all time steps $t = 1, 2, \ldots, T$ 
and only then forwards the accumulated results to layer $\ell+1$.

\paragraph{Recommendation for RMPA.}
Even when RMPA is deployed without its hardware-level optimizations, we recommend 
adopting multi-step mode for the following reasons:

\begin{itemize}[leftmargin=*, nosep]
    \item \textbf{Hardware compatibility.} Multi-step mode aligns more naturally with 
    the dataflow patterns of contemporary hardware accelerators, where minimizing 
    memory access overhead is critical for sustained throughput.

    \item \textbf{Reduced weight reloading.} By completing all time steps within a 
    given layer before advancing to the next, multi-step mode eliminates the need to 
    repeatedly reload the same synaptic weights for each time step, thereby 
    significantly reducing memory bandwidth pressure.

    \item \textbf{Improved parallelism.} Multi-step mode naturally exposes 
    intra-layer parallelism across time steps, enabling more efficient utilization of 
    vectorized or batched computation resources.
\end{itemize}

\paragraph{Adapting RMPA for Single-Step Mode.}
If single-step mode is required by the deployment scenario, a minor modification to 
RMPA is necessary. Specifically, for the second and all subsequent time steps 
($t \geq 2$), when newly generated spike-driven data is forwarded to downstream 
layers, the membrane potential $m_\ell$ of layer $\ell$ must be updated by incorporating 
the incoming contribution scaled by $2^{i-1}$:
\begin{equation}
    \mathbf{m}_\ell(t) \leftarrow V_\ell(t) + 2^{i-1} \cdot \mathbf{I}_\ell(t), \qquad t \geq 2,
\end{equation}
where $\mathbf{I}_\ell(t)$ denotes the input increment arriving at layer $\ell$ at time step $t$, 
and $i$ is the number of RMPA loop. This scaling ensures that the accumulated membrane 
potential correctly reflects the appropriately amplified value.

\section{Additional Experimental Results When Qeury Converted}
For the spiking conversion of Transformer models, this paper investigates the trade-off between computational efficiency and model accuracy in the treatment of the query (q) representation. By default, 
q is retained in floating-point form and excluded from spike conversion, while spiking conversion is applied exclusively to the key (k) and value(v). This design treats q as a weight-like parameter, which eliminates floating-point multiply-accumulate operations in the attention mechanism while preserving model accuracy. To assess the impact of this design choice, we conduct ablation experiments on whether to convert q into spikes, with results reported in Table~\ref{tab:query}.

\begin{table*}[htbp]
  \centering
  \caption{Experiments on ViT models under different $(T,D)$ settings. \texttt{q\_include} denotes the variant that also converts the query ($q$) into spikes (in addition to $k$ and $v$), while \texttt{Default} converts only $k$ and $v$.}
  \label{tab:query}
  \resizebox{\textwidth}{!}{
  \begin{tabular}{llcccccccccc}
    \toprule
    \multicolumn{1}{c}{\multirow{2}{*}{Task}} & 
    \multicolumn{1}{c}{\multirow{2}{*}{Model}} & 
    \multicolumn{1}{c}{\multirow{2}{*}{ANN}} & 
    \multicolumn{1}{c}{\multirow{2}{*}{$D$}} & 
    \multicolumn{3}{c}{Default} & 
    \multicolumn{3}{c}{q\_include} \\
    \cmidrule(lr){5-7} \cmidrule(lr){8-10}
    & & & & $T{=}1$ & $T{=}2$ & $T{=}3$ & $T{=}1$ & $T{=}2$ & $T{=}3$ \\
    
    \midrule
    \multirow{10}{*}{ImageNet-1K} 
    & \multirow{3}{*}{vit\_small\_patch16\_224} & \multirow{3}{*}{78.04}
    & 1 & 48.20 & 76.27 & 77.80 & 42.34 & 76.25 & 77.87 \\
    & & & 2 & 73.78 & 77.94 & 78.03 & 73.26 & 77.93 & 78.04 \\
    & & & 3 & 77.01 & 78.06 & 78.03 & 76.94 & 78.06 & 78.03 \\
    \cmidrule(lr){2-10}
    & \multirow{3}{*}{vit\_base\_patch16\_224} & \multirow{3}{*}{82.91}
    & 1 & 74.00 & 82.21 & 82.74 & 72.37 & 82.17 & 82.73 \\
    & & & 2 & 80.89 & 82.77 & 82.89 & 80.71 & 82.76 & 82.87 \\
    & & & 3 & 82.41 & 82.87 & 82.91 & 82.43 & 82.87 & 82.92 \\
    \cmidrule(lr){2-10}
    & \multirow{3}{*}{vit\_large\_patch16\_224} & \multirow{3}{*}{84.88}
    & 1 & 80.89 & 84.46 & 84.79 & 80.26 & 84.43 & 84.79 \\
    & & & 2 & 84.03 & 84.80 & 84.89 & 83.96 & 84.81 & 84.89 \\
    & & & 3 & 84.45 & 84.89 & 84.88 & 84.41 & 84.89 & 84.88 \\
    
    \midrule
    
    \multirow{15}{*}{MMLU}
    & \multirow{3}{*}{qwen3\_0.6b\_base} & \multirow{3}{*}{52.41}
    & 1 & 29.06 & 50.29 & 52.11 & 25.58 & 49.47 & 52.31 \\
    & & & 2 & 45.85 & 52.55 & 52.56 & 43.40 & 52.24 & 52.73 \\
    & & & 3 & 51.08 & 52.60 & 52.49 & 50.11 & 52.60 & 52.60 \\
    \cmidrule(lr){2-10}
    & \multirow{3}{*}{qwen3\_1.7b\_base} & \multirow{3}{*}{62.61}
    & 1 & 41.59 & 60.80 & 62.37 & 31.94 & 60.12 & 62.47 \\
    & & & 2 & 57.83 & 62.42 & 62.53 & 55.62 & 62.34 & 62.78 \\
    & & & 3 & 60.87 & 62.26 & 62.50 & 60.93 & 62.63 & 62.56 \\
    \cmidrule(lr){2-10}
    & \multirow{3}{*}{qwen3\_4b\_base} & \multirow{3}{*}{72.90}
    & 1 & 57.58 & 72.34 & 72.70 & 43.85 & 71.57 & 72.72 \\
    & & & 2 & 69.29 & 72.75 & 73.10 & 67.40 & 72.62 & 73.07 \\
    & & & 3 & 72.21 & 72.92 & 73.04 & 71.85 & 73.00 & 73.04 \\
    \cmidrule(lr){2-10}
    & \multirow{3}{*}{qwen3\_8b\_base} & \multirow{3}{*}{76.85}
    & 1 & 65.26 & 76.13 & 76.86 & 54.42 & 75.97 & 76.51 \\
    & & & 2 & 74.88 & 76.71 & 76.78 & 73.06 & 76.59 & 76.81 \\
    & & & 3 & 76.45 & 76.90 & 76.81 & 75.99 & 76.81 & 76.73 \\
    \cmidrule(lr){2-10}
    & \multirow{3}{*}{qwen3\_14b\_base} & \multirow{3}{*}{80.47}
    & 1 & 70.70 & 79.73 & 80.39 & 62.08 & 79.58 & 80.37 \\
    & & & 2 & 78.63 & 80.47 & 80.45 & 77.43 & 80.42 & 80.65 \\
    & & & 3 & 80.37 & 80.49 & 80.49 & 79.83 & 80.40 & 80.44 \\

    \bottomrule
  \end{tabular}
  }
\end{table*}

\section{Additional Experimental Results on ImageNet-1K}\label{overall_imagenet}

This section presents comprehensive experimental results of our SNN conversion framework on the ImageNet-1K dataset across various models, including EVA02, ViT, ResNet, and VGG. We report accuracy in different timesteps \(T\) and spike dimensions \(D\) in table~\ref{tab:appendix_imagenet}. The ANN baseline accuracy is provided for reference.

\begin{table*}[htbp]
\centering
\caption{Accuracy (\%) on ImageNet-1K Dataset under Different $T$ and $D$ Settings}
\label{tab:appendix_imagenet}
\begin{minipage}{0.48\textwidth}
\centering
\begin{tabular}{lcccc}
\toprule
\textbf{Model} & $T=1$ & $T=2$ & $T=3$ & \textbf{ANN} \\
\midrule
\multicolumn{5}{l}{\textbf{eva02\_tiny\_patch14\_336}} \\
$d=1$ & 73.26 & 80.11 & 80.53 & 80.62 \\
$d=2$ & 78.82 & 80.60 & 80.65 \\
$d=3$ & 80.19 & 80.63 & 80.61 \\
\midrule
\multicolumn{5}{l}{\textbf{eva02\_small\_patch14\_336}} \\
$d=1$ & 80.18 & 85.22 & 85.68 & 85.73 \\
$d=2$ & 84.54 & 85.74 & 85.75 \\
$d=3$ & 85.45 & 85.70 & 85.73 \\
\midrule
\multicolumn{5}{l}{\textbf{eva02\_base\_patch14\_448}} \\
$d=1$ & 81.53 & 87.85 & 88.25 & 88.26 \\
$d=2$ & 86.88 & 88.28 & 88.26 \\
$d=3$ & 88.00 & 88.27 & 88.26 \\
\midrule
\multicolumn{5}{l}{\textbf{eva02\_large\_patch14\_448}} \\
$d=1$ & 85.66 & 89.32 & 89.57 & 89.63 \\
$d=2$ & 88.88 & 89.60 & 89.63 \\
$d=3$ & 89.46 & 89.63 & 89.63 \\
\midrule
\multicolumn{5}{l}{\textbf{eva\_giant\_patch14\_336}} \\
$d=1$ & 88.65 & 89.41 & 89.57 & 89.62 \\
$d=2$ & 89.27 & 89.56 & 89.60 \\
$d=3$ & 89.47 & 89.58 & 89.62 \\
\bottomrule
\end{tabular}
\end{minipage}
\hfill
\begin{minipage}{0.48\textwidth}
\centering
\begin{tabular}{lcccc}
\toprule
\textbf{Model} & $T=1$ & $T=2$ & $T=3$ & \textbf{ANN} \\
\midrule
\multicolumn{5}{l}{\textbf{vit\_small\_patch16\_224}} \\
$d=1$ & 48.20 & 76.27 & 77.80 & 78.04 \\
$d=2$ & 73.78 & 77.94 & 78.03 \\
$d=3$ & 77.01 & 78.06 & 78.03 \\
\midrule
\multicolumn{5}{l}{\textbf{vit\_base\_patch16\_224}} \\
$d=1$ & 74.00 & 82.21 & 82.74 & 82.91 \\
$d=2$ & 80.89 & 82.77 & 82.89 \\
$d=3$ & 82.41 & 82.87 & 82.91 \\
\midrule
\multicolumn{5}{l}{\textbf{vit\_large\_patch16\_224}} \\
$d=1$ & 80.89 & 84.46 & 84.79 & 84.88 \\
$d=2$ & 84.03 & 84.80 & 84.89 \\
$d=3$ & 84.45 & 84.89 & 84.88 \\
\midrule
\multicolumn{5}{l}{\textbf{resnet34}} \\
$d=1$ & 20.23 & 72.45 & 75.83 & 76.09 \\
$d=2$ & 66.59 & 76.08 & 76.12 \\
$d=3$ & 74.81 & 76.13 & 76.09 \\
\midrule
\multicolumn{5}{l}{\textbf{vgg16}} \\
$d=1$ & 49.82 & 68.86 & 70.97 & 71.12 \\
$d=2$ & 65.25 & 70.94 & 71.10 \\
$d=3$ & 69.72 & 71.08 & 71.12 \\
\bottomrule
\end{tabular}
\end{minipage}
\end{table*}

\section{Experimental Results on CIFAR datasets}\label{overall_cifars}


\begin{figure}[ht!]
    \centering
    \includegraphics[width=1\linewidth]{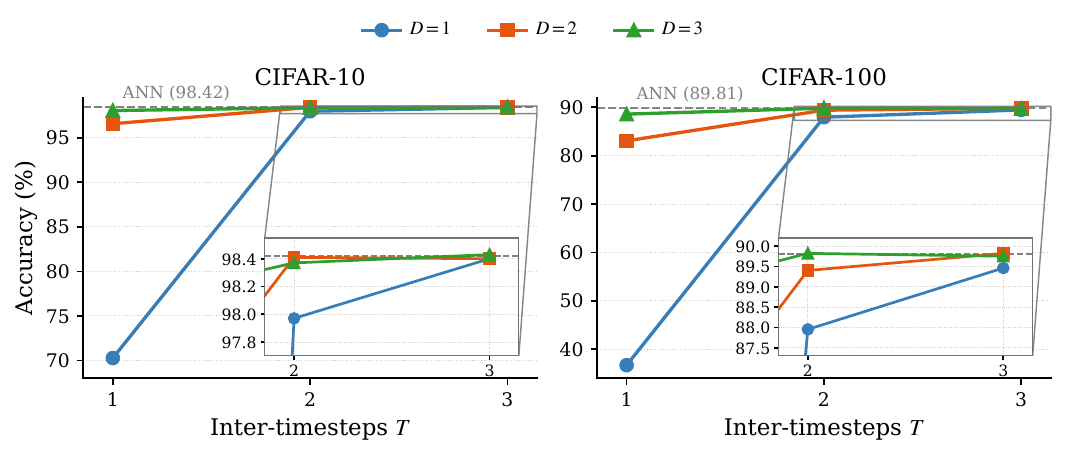}
    \caption{Test accuracy of ViT-S on CIFAR-10 (left) and CIFAR-100 (right) under different settings of  $T$ and $D$.}
    \label{fig:cifar}
\end{figure}

We evaluate the converted ViT-S model in CIFAR tasks under different combinations of parameters and report the post-conversion accuracy in figure~\ref{fig:cifar}.
For CIFAR-10, with \(D=1\), our method achieves \textbf{98.40\%} accuracy at \(T=3\), closely matching the ANN baseline (98.42\%) while requiring only 3 timesteps. Remarkably, even at \(T=2\), the accuracy reaches 97.97\%, with a negligible drop of 0.45\%. For CIFAR-100, our method attains \textbf{89.46\%} at \(T=3\) and 87.95\% at \(T=2\), again approaching the baseline of the ANN of 89.81\%.

We further observe that deeper networks enable faster convergence. With \(D=3\), the SNN on CIFAR-10 already reaches 98.05\% at \(T=1\) and saturates at 98.37\% for \(T=3\), effectively matching the ANN. Similarly, on CIFAR-100, \(D=3\) yields 88.59\% at \(T=1\) and 89.82\% at \(T=2\), narrowing the gap to within 0.06\% of the ANN. These results demonstrate that our method achieves almost lossless conversion with as few as \(T=2\) timesteps, and deeper architectures further reduce the required latency. 


\section{Experimental Results on MMLU Benchmark}\label{overall_mmlu}

\begin{figure}[ht!]
    \centering
    \includegraphics[width=0.7\linewidth]{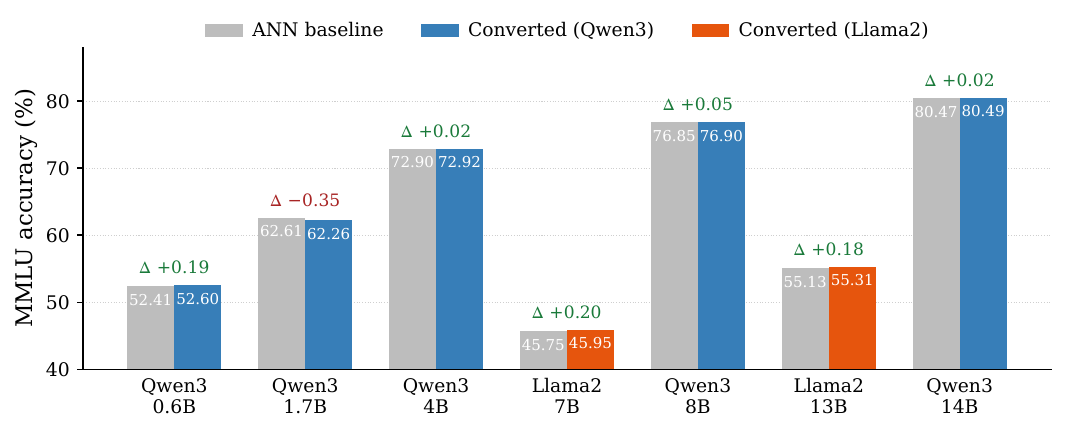}
    \caption{Conversion results on MMLU for different models}
    \label{fig:mmlu}
\end{figure}

The Qwen3\cite{qwen3} and Llama2\cite{llama2} series models are evaluated on the  MMLU benchmark in this work. We perform comprehensive evaluations on this key NLP benchmark, and the corresponding results are shown in figure~\ref{fig:mmlu}.

Among the 29 evaluation items, only the Llama2-7B model on the Winogrande task and the Qwen3-1.7B-Base model on the MMLU task suffer accuracy drops of 0.32\% and 0.35\%, respectively. Meanwhile, the Llama2-7B model incurs accuracy losses of 0.12\% and 0.1\% in the ARC-E and HellaSwag\_norm tasks, respectively. For the remaining 25 tasks, the accuracy loss is less than 0.1\%, which verifies the universality and effectiveness of the proposed paradigm in CV and NLP tasks.
\section{Additional Experimental Results of LLaMA2-7B and LLaMA2-13b}\label{overall_llama}

This section presents comprehensive experimental results of our SNN conversion framework on various NLP tasks for the LLaMA2-7B and LLaMA2-13B models in table~\ref{tab:appendix_llama}. We report accuracy in different timesteps \(T\) and spike dimensions \(D\). The ANN baseline accuracy is provided for reference.

\begin{table*}[htbp!]
\centering
\caption{Accuracy (\%) of LLaMA2-7B and LLaMA2-13B on NLP Tasks under Different $T$ and $D$ Settings}
\label{tab:appendix_llama}
\resizebox{0.8\textwidth}{!}{
\begin{tabular}{lccccclcccc}
\toprule
& \multicolumn{4}{c}{\textbf{LLaMA2-7B}} & & \multicolumn{4}{c}{\textbf{LLaMA2-13B}} \\
\cmidrule(lr){2-5} \cmidrule(lr){7-10}
\textbf{Task} & $$T{=}1$$ & $$T{=}2$$ & $$T{=}3$$ & ANN & & $$T{=}1$$ & $$T{=}2$$ & $$T{=}3$$ & ANN \\
\midrule
\multicolumn{10}{l}{\textbf{MMLU}} \\
$$d{=}1$$ & 26.90 & 43.88 & 45.93 & 45.75 & & 33.76 & 53.62 & 54.99 & 55.13 \\
$$d{=}2$$ & 39.97 & 45.67 & 45.73 &  & & 51.40 & 55.05 & 55.23 &  \\
$$d{=}3$$ & 44.92 & 45.95 & 45.93 &  & & 54.15 & 55.31 & 55.14 & \\
\midrule
\multicolumn{10}{l}{\textbf{PIQA-norm}} \\
$$d{=}1$$ & 72.14 & 79.11 & 78.56 & 78.45 & & 74.32 & 80.09 & 80.52 & 80.52 \\
$$d{=}2$$ & 77.37 & 78.73 & 78.84 &  & & 79.60 & 80.69 & 80.25 &  \\
$$d{=}3$$ & 78.78 & 78.67 & 78.51 & & & 80.25 & 80.58 & 80.47 &  \\
\midrule
\multicolumn{10}{l}{\textbf{ARC-Challenge norm}} \\
$$d{=}1$$ & 33.19 & 44.88 & 45.48 & 45.73 & & 39.76 & 48.55 & 48.55 & 48.89 \\
$$d{=}2$$ & 43.60 & 45.48 & 45.65 &  & & 47.78 & 48.72 & 48.38 &  \\
$$d{=}3$$ & 46.16 & 45.65 & 45.99 &  & & 49.40 & 49.15 & 48.55 & \\
\midrule
\multicolumn{10}{l}{\textbf{ARC-Easy norm}} \\
$$d{=}1$$ & 58.33 & 73.61 & 74.75 & 74.87 & & 66.79 & 75.93 & 76.60 & 76.56 \\
$$d{=}2$$ & 72.26 & 73.99 & 74.54 &  & & 75.51 & 76.26 & 76.47 &  \\
$$d{=}3$$ & 73.82 & 74.75 & 74.66 &  & & 75.80 & 76.47 & 76.43 & \\
\midrule
\multicolumn{10}{l}{\textbf{Winogrande norm}} \\
$$d{=}1$$ & 60.38 & 69.38 & 69.46 & 69.38 & & 62.67 & 71.90 & 71.90 & 72.22 \\
$$d{=}2$$ & 67.01 & 69.53 & 69.30 & & & 71.43 & 72.53 & 72.22 &  \\
$$d{=}3$$ & 67.56 & 69.06 & 69.46 &  & & 70.56 & 72.14 & 72.53 &  \\
\midrule
\multicolumn{10}{l}{\textbf{HellaSwag norm}} \\
$$d{=}1$$ & 65.61 & 75.72 & 76.30 & 76.28 & & 66.05 & 79.04 & 79.55 & 79.71 \\
$$d{=}2$$ & 74.65 & 76.25 & 76.18 &  & & 78.57 & 79.63 & 79.71 &  \\
$$d{=}3$$ & 75.77 & 76.18 & 76.25 &  & & 79.23 & 79.69 & 79.72 & \\
\bottomrule
\end{tabular}
}
\end{table*}

\section{MMLU Results of Qwen3 Series Models}
This section supplements the accuracy of the MMLU task of the Qwen3 series with different timesteps \(T\) and spike dimensions \(D\) in table~\ref{tab:appendix_qwen3_mmlu}, with the original ANN accuracy provided as the baseline reference.

\begin{table*}[htbp!]
\centering
\caption{MMLU Accuracy (\%) of Qwen3 Series Models under Different $T$ and $D$ Settings}
\label{tab:appendix_qwen3_mmlu}
\resizebox{0.55\textwidth}{!}{
\begin{tabular}{lcccc}
\toprule
\textbf{Model} & $$T=1$$ & $$T=2$$ & $$T=3$$ & \textbf{ANN} \\
\midrule
\multicolumn{5}{l}{\textbf{qwen3-0.6b-base}} \\
$$d=1$$ & 29.06 & 50.29 & 52.11 & 52.41 \\
$$d=2$$ & 45.85 & 52.55 & 52.56 & \\
$$d=3$$ & 51.08 & 52.60 & 52.49 & \\
\midrule
\multicolumn{5}{l}{\textbf{qwen3-1.7b-base}} \\
$$d=1$$ & 41.59 & 60.80 & 62.37 & 62.61 \\
$$d=2$$ & 57.83 & 62.42 & 62.53 & \\
$$d=3$$ & 60.87 & 62.26 & 62.50 & \\
\midrule
\multicolumn{5}{l}{\textbf{qwen3-4b-base}} \\
$$d=1$$ & 57.58 & 72.34 & 72.70 & 72.90 \\
$$d=2$$ & 69.29 & 72.75 & 73.10 &  \\
$$d=3$$ & 72.21 & 72.92 & 73.04 &  \\
\midrule
\multicolumn{5}{l}{\textbf{qwen3-8b-base}} \\
$$d=1$$ & 65.26 & 76.13 & 76.86 & 76.85 \\
$$d=2$$ & 74.88 & 76.71 & 76.78 &  \\
$$d=3$$ & 76.45 & 76.90 & 76.81 &  \\
\midrule
\multicolumn{5}{l}{\textbf{qwen3-14b-base}} \\
$$d=1$$ & 70.70 & 79.73 & 80.39 & 80.47 \\
$$d=2$$ & 78.63 & 80.47 & 80.45 &  \\
$$d=3$$ & 80.37 & 80.49 & 80.49 &  \\
\bottomrule
\end{tabular}
}
\end{table*}

\clearpage
\section*{NeurIPS Paper Checklist}

The checklist is designed to encourage best practices for responsible machine learning research, addressing issues of reproducibility, transparency, research ethics, and societal impact. Do not remove the checklist: {\bf The papers not including the checklist will be desk rejected.} The checklist should follow the references and follow the (optional) supplemental material.  The checklist does NOT count towards the page
limit. 

Please read the checklist guidelines carefully for information on how to answer these questions. For each question in the checklist:
\begin{itemize}
    \item You should answer \answerYes{}, \answerNo{}, or \answerNA{}.
    \item \answerNA{} means either that the question is Not Applicable for that particular paper or the relevant information is Not Available.
    \item Please provide a short (1--2 sentence) justification right after your answer (even for \answerNA). 
\end{itemize}

{\bf The checklist answers are an integral part of your paper submission.} They are visible to the reviewers, area chairs, senior area chairs, and ethics reviewers. You will also be asked to include it (after eventual revisions) with the final version of your paper, and its final version will be published with the paper.

The reviewers of your paper will be asked to use the checklist as one of the factors in their evaluation. While \answerYes{} is generally preferable to \answerNo{}, it is perfectly acceptable to answer \answerNo{} provided a proper justification is given (e.g., error bars are not reported because it would be too computationally expensive'' or ``we were unable to find the license for the dataset we used''). In general, answering \answerNo{} or \answerNA{} is not grounds for rejection. While the questions are phrased in a binary way, we acknowledge that the true answer is often more nuanced, so please just use your best judgment and write a justification to elaborate. All supporting evidence can appear either in the main paper or the supplemental material, provided in appendix. If you answer \answerYes{} to a question, in the justification please point to the section(s) where related material for the question can be found.



\begin{enumerate}

\item {\bf Claims}
    \item[] Question: Do the main claims made in the abstract and introduction accurately reflect the paper's contributions and scope?
    \item[] Answer: \answerYes{}
    \item[] Justification: The abstract and introduction clearly state the main contributions, including the PRIME framework, RMPA mechanism, hardware-friendly inference scheme, and experimental results on CV and NLP tasks.

\item {\bf Limitations}
    \item[] Question: Does the paper discuss the limitations of the work performed by the authors?
    \item[] Answer: \answerYes{}
    \item[] Justification: In Section 5, we discuss the limitations of this work: as with most algorithm-level works, the efficiency figures reported here are based on analytical estimation rather than direct on-chip measurement. However, given that all underlying computational primitives (bit-shift, integer addition, leading-one detection) are natively supported by standard digital logic, physical implementation remains a natural and straightforward next step.

\item {\bf Theory assumptions and proofs}
    \item[] Question: For each theoretical result, does the paper provide the full set of assumptions and a complete (and correct) proof?
    \item[] Answer: \answerYes{}
    \item[] Justification: Complete theoretical error bounds for both conventional rate coding and the proposed integration coding are provided in the Error Analysis section. See Appendix A.

    \item {\bf Experimental result reproducibility}
    \item[] Question: Does the paper fully disclose all the information needed to reproduce the main experimental results of the paper to the extent that it affects the main claims and/or conclusions of the paper (regardless of whether the code and data are provided or not)?
    \item[] Answer: \answerYes{}
    \item[] Justification: Sufficient experimental details, including datasets, model architectures, and evaluation configurations, are provided in the experimental section and the appendix to allow reproduction of the main results. The source code will be made publicly available upon acceptance.

\item {\bf Open access to data and code}
    \item[] Question: Does the paper provide open access to the data and code, with sufficient instructions to faithfully reproduce the main experimental results, as described in supplemental material?
     \item[] Answer: \answerYes{}
    \item[] Justification: Code will be made publicly available upon acceptance. All datasets used are standard public benchmarks.

\item {\bf Experimental setting/details}
    \item[] Question: Does the paper specify all the training and test details (e.g., data splits, hyperparameters, how they were chosen, type of optimizer) necessary to understand the results?
    \item[] Answer: \answerYes{}
    \item[] Justification: This work proposes a training-free ANN-to-SNN conversion framework. Inference configurations and evaluation settings are fully described in the experimental section.

\item {\bf Experiment statistical significance}
    \item[] Question: Does the paper report error bars suitably and correctly defined or other appropriate information about the statistical significance of the experiments?
    \item[] Answer: \answerNo{}
    \item[] Justification: Error bars are not reported. The proposed method is deterministic given a fixed pre-trained ANN, and result variance across runs is negligible.

\item {\bf Experiments compute resources}
    \item[] Question: For each experiment, does the paper provide sufficient information on the computer resources (type of compute workers, memory, time of execution) needed to reproduce the experiments?
     \item[] Answer: \answerYes{}
    \item[] Justification: Although this work focuses on ANN-to-SNN conversion methodology, we acknowledge that inference accuracy may exhibit negligible deviations due to variations in hardware devices. Accordingly, we report the inference devices used in our experiments.
    
\item {\bf Code of ethics}
    \item[] Question: Does the research conducted in the paper conform, in every respect, with the NeurIPS Code of Ethics \url{https://neurips.cc/public/EthicsGuidelines}?
    \item[] Answer: \answerYes{}
    \item[] Justification: This work addresses energy-efficient neural network inference and raises no ethical concerns.

\item {\bf Broader impacts}
    \item[] Question: Does the paper discuss both potential positive societal impacts and negative societal impacts of the work performed?
   \item[] Answer: \answerNA{}
    \item[] Justification: This is foundational hardware-efficiency research with no foreseeable direct negative societal impact.
    
\item {\bf Safeguards}
    \item[] Question: Does the paper describe safeguards that have been put in place for responsible release of data or models that have a high risk for misuse (e.g., pre-trained language models, image generators, or scraped datasets)?
    \item[] Answer: \answerNA{}
    \item[] Justification: The paper does not release high-risk models or datasets.

\item {\bf Licenses for existing assets}
    \item[] Question: Are the creators or original owners of assets (e.g., code, data, models), used in the paper, properly credited and are the license and terms of use explicitly mentioned and properly respected?
    \item[] Answer: \answerYes{}
    \item[] Justification: All datasets and code libraries used in this paper are properly cited, and their licenses are acknowledged.

\item {\bf New assets}
    \item[] Question: Are new assets introduced in the paper well documented and is the documentation provided alongside the assets?
     \item[] Answer: \answerNA{}
    \item[] Justification: No new datasets, model weights, or benchmarks are released in this paper.

\item {\bf Crowdsourcing and research with human subjects}
    \item[] Question: For crowdsourcing experiments and research with human subjects, does the paper include the full text of instructions given to participants and screenshots, if applicable, as well as details about compensation (if any)? 
    \item[] Answer: \answerNA{}
    \item[] Justification: This paper involves no crowdsourcing or human subjects research.

\item {\bf Institutional review board (IRB) approvals or equivalent for research with human subjects}
    \item[] Question: Does the paper describe potential risks incurred by study participants, whether such risks were disclosed to the subjects, and whether Institutional Review Board (IRB) approvals (or an equivalent approval/review based on the requirements of your country or institution) were obtained?
    \item[] Answer: \answerNA{}
    \item[] Justification: This paper involves no human subjects research.

\item {\bf Declaration of LLM usage}
    \item[] Question: Does the paper describe the usage of LLMs if it is an important, original, or non-standard component of the core methods in this research? Note that if the LLM is used only for writing, editing, or formatting purposes and does \emph{not} impact the core methodology, scientific rigor, or originality of the research, declaration is not required.
    \item[] Answer: \answerNA{}
    \item[] Justification: LLMs (e.g., LLaMA) appear in this paper solely as evaluation targets for the proposed conversion framework, not as components of the core methodology.

\end{enumerate}

\end{document}